\documentclass{article}

\usepackage{microtype}
\usepackage{graphicx}
\usepackage{subfigure}
\usepackage{booktabs} 

\usepackage{hyperref}

\newcommand{\theHalgorithm}{\arabic{algorithm}}

\usepackage[accepted]{icml2024}

\usepackage{amsmath}
\usepackage{amssymb}
\usepackage{mathtools}
\usepackage{amsthm}
\usepackage{stmaryrd}
\SetSymbolFont{stmry}{bold}{U}{stmry}{m}{n}
\usepackage{xspace}

\usepackage[capitalize,noabbrev]{cleveref}

\usepackage{pifont}
\newcommand{\cmark}{\ding{51}}
\newcommand{\xmark}{\ding{55}}

\usepackage{multirow,multicol}

\theoremstyle{plain}

\theoremstyle{definition}

\theoremstyle{remark}

\usepackage{xargs}
\usepackage[colorinlistoftodos,prependcaption,textsize=tiny]{todonotes}

\icmltitlerunning{Fine-grained Classes and How to Find Them}

\begin{document}

\twocolumn[
\icmltitle{Fine-grained Classes and How to Find Them}



\icmlsetsymbol{equal}{*}

\begin{icmlauthorlist}
\icmlauthor{Matej Grcić}{equal,epfl,fer}
\icmlauthor{Artyom Gadetsky}{equal,epfl}
\icmlauthor{Maria Brbić}{epfl}
\end{icmlauthorlist}

\icmlaffiliation{fer}{Faculty of Electrical Engineering and Computing, University of Zagreb, Croatia}
\icmlaffiliation{epfl}{EPFL, Lausanne, Switzerland}

\icmlcorrespondingauthor{Maria Brbić}{mbrbic@epfl.ch}

\icmlkeywords{Fine-grained class discovery, Fine-grained classification, Discrete optimization, Class relationships, Machine Learning, ICML}

\vskip 0.3in
]



\printAffiliationsAndNotice{\icmlEqualContribution} 

\begin{abstract}
In many practical applications, coarse-grained labels are readily available compared to fine-grained labels that reflect subtle differences between classes. However, existing methods cannot leverage coarse labels to infer fine-grained labels in an unsupervised manner. To bridge this gap, we propose FALCON, a method that discovers fine-grained classes from coarsely labeled data without any supervision at the fine-grained level. FALCON simultaneously infers unknown fine-grained classes and underlying relationships between coarse and fine-grained classes. Moreover, FALCON is a modular method that can effectively learn from multiple datasets labeled with different strategies. We evaluate FALCON on eight image classification tasks and a single-cell classification task. FALCON outperforms baselines by a large margin, achieving  $22\%$  improvement over the best baseline on the tieredImageNet dataset with over $600$ fine-grained classes.
\end{abstract}

\section{Introduction}

Machine learning excels in domains with large quantities of precisely labeled data \cite{esteva17n,kirillov23iccv}. While coarse labels are typically abundant and easy to obtain, precise annotation with fine-grained labels is challenging due to the subtle differences between classes and the small number of discriminative features. Thus, in many domains obtaining such fine-grained labels requires domain expertise and tedious manual effort \cite{tkatchenko20nc,erfanian23bp}. For example,  B-cells and T-cells can be easily differentiated, but differentiating between very fine-grained cell subtypes such as CD4+ T cells and CD8+ T cells requires identifying a very small number of specific markers. To automate the tedious effort of obtaining fine-grained labels, machine learning methods that can differentiate between subtle differences in fine-grained labels are needed.

Prior work has shown that coarse labels can be used to more effectively learn fine-grained classes \cite{wu18eccv}.
Weakly-supervised classification methods use coarse labels as a form of weak supervision to improve fine-grained classification performance \cite{ristin15cvpr, taherkhani19iccv}. Recently, few-shot learning methods have been proposed that are trained on a set of coarse classes and then adapted for fine-grained classification with only a few labeled samples per class \cite{liu19ijcai,bukchin21cvpr,ni22iclr}. However, all these methods assume that a set of fine-grained classes along with a small number of samples assigned to them are known beforehand. 

In this work, we propose FALCON (\textbf{F}ine gr\textbf{A}ined \textbf{L}abels from \textbf{CO}arse supervisio\textbf{N}), a method that discovers fine-grained classes within a coarsely labeled dataset without any supervision. The key insight in FALCON is that fine-grained predictions can be combined with the relations between coarse and fine classes to recover coarse predictions. With this insight, FALCON develops a specialized optimization procedure that alternates between inferring unknown relations between the coarse and fine-grained classes and training a fine-grained classifier. Relationships between the coarse and fine-grained classes are inferred by solving a
discrete optimization problem, while the fine-grained classifier is trained using coarse supervision and fine-grained pseudo-labels. Moreover, FALCON can be seamlessly adapted to leverage multiple datasets with incompatible coarse classes, relabeling all of them at the same fine-grained level.


We compare FALCON to alternative baselines on eight image classification datasets and a single-cell dataset from biology domain. Experimental results show that FALCON effectively discovers fine-grained classes without supervision and consistently outperforms baselines on both image and single-cell data. For instance, on the tieredImageNet dataset with $608$ fine-grained classes, FALCON outperforms baselines by $22\%$. Moreover, when trained with multiple datasets with different coarse classes, FALCON effectively reuses different annotation policies to improve its performance.






\section{Related Work}

\textbf{Weakly-supervised classification.}
Coarse labels can be utilized as a form of weak supervision. Previous works \citep{ristin15cvpr,taherkhani19iccv,hsieh19arxiv} boost fine-grained classification performance by training on a mixture of coarse and fine labels. \citet{robinson20icml} build a theoretical framework for analyzing coarse labels as a form of weak supervision.
However, all these methods are trained in a supervised manner with both coarse and fine supervision. In contrast, FALCON does not use any fine-grained supervision and assumes that only coarse labels are available during training.

\textbf{Cross-granularity few-shot learning.} Cross-granularity few-shot learning \citep{ni22iclr} considers a setting where a model is trained on a set
of coarse classes and it needs to adapt to the fine-grained classes given only a few labeled examples. 
\citet{wu18eccv} combines a non-parametric classifier with a deep feature extractor to learn fine-grained representation space. \citet{bukchin21cvpr} propose a contrastive learning objective that decreases the angle between the augmented views of the same input while increasing the angle to other samples from the same coarse class.
\citet{ni22iclr} models the generation of samples from coarse classes by representing the fine-grained subclasses with latent variables.
\citet{liu19ijcai} considers a few-shot setup where relations between coarse and fine classes are known beforehand.
Our setting differs from the few-shot learning setting since we do not assume to have any fine-grained labels available. Instead, our goal is to discover fine-grained classes.

\textbf{Fine-grained image retrieval.}
Fine-grained image retrieval \citep{touvron21iccv} utilizes coarse supervision to learn fine-grained representations.
The learned representations  capture fine-grained correspondences between samples and enable retrieval of images at finer granularity.
To solve this task, previous works \cite{xu21iccv,touvron21iccv} combine self-supervised representation learning with parametric or non-parametric coarse classifiers.
Alternatively, \citet{feng23cvpr} extends the self-supervised objective by reweighting the samples according to the similarity in the feature space. 
Different from image retrieval, our goal is to discover the underlying fine-grained classes within the coarsely labeled data.

\textbf{Deep clustering.} 
Clustering is a decades-old machine learning problem \cite{lance67cj}. 
Recent deep clustering methods \cite{chang17iccv,asano20iclr, gadetsky2023pursuit} outperform the classical approaches using deep neural networks trained with carefully designed optimization objectives.
Common deep clustering objectives encourage consistent predictions for similar samples \cite{gansbeke20eccv}, or gradually fit the network to the most confident predictions \cite{niu22tip,asano20iclr}. 
Deep clustering methods can be used for the inference of fine-grained classes in an unsupervised manner; however, they typically result in a suboptimal performance due to the difficulty of the task. To address this limitation, we design a method that can effectively utilize available coarse-level supervision for the discovery of fine-level classes.


\section{Fine-grained Class Discovery}
\label{sec:method}

\begin{figure*}[ht]
\begin{center}
\centerline{\includegraphics[width=\linewidth]{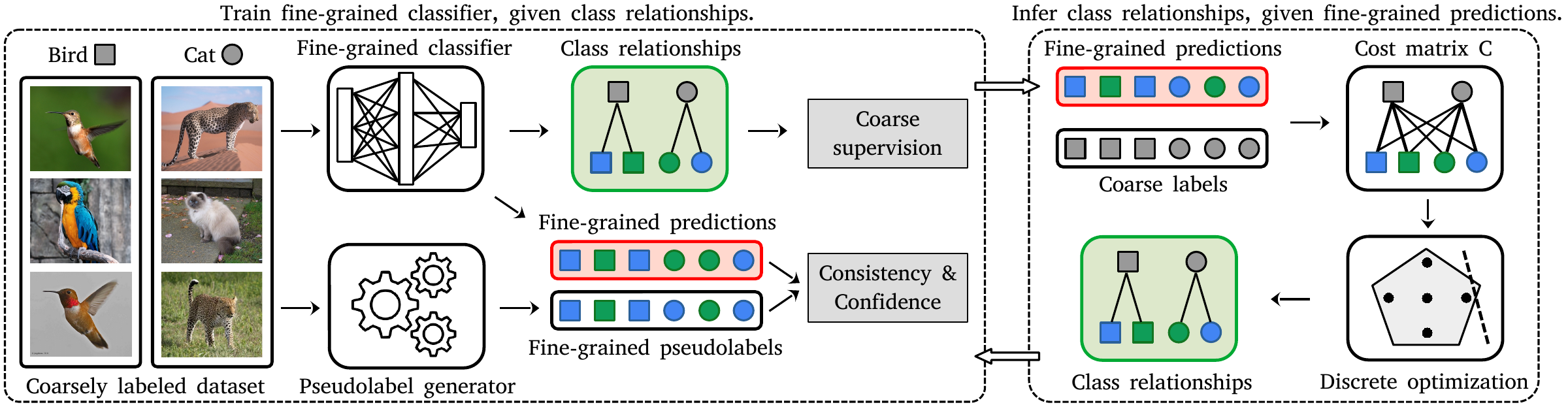}}
\vskip -0.1in
\caption{
FALCON simultaneously discovers fine-grained classes and infers relationships between the discovered fine and the available coarse classes by coarse supervision.
The fine-grained classifier optimizes the loss (\ref{eq:final_cls}), while the class relationships are inferred by solving a discrete optimization problem (\ref{eq:objective_M}).
}
\label{fig:task_c2f}
\end{center}
\vskip -0.35in
\end{figure*}

\textbf{Problem setup.} Let $\mathcal{X}$ be a sample space and $\mathcal{Y}_C$ a set of $K_C$ coarse classes.
We assume a coarsely labeled dataset $\mathcal{D} = \{(\mathbf{x}^i, y_c^i)\}_{i=1}^N$ is given,
where $\mathbf{x}^i \in \mathcal{X}$ and $ y_c^i \in \mathcal{Y}_C$.
Additionally, every sample $\mathbf{x} \in \mathcal{D} $ is associated with a fine-grained class $y_\text{f}$ from an unknown set of fine-grained classes $\mathcal{Y}_F$.
We assume that every fine-grained class $y_\text{f} \in \mathcal{Y}_F$ is associated with a single coarse class $y_c \in \mathcal{Y}_C$, \textit{i.e.}, has a single coarse-grained parent. The number of fine classes $K_F = |\mathcal{Y}_F|$ is greater than $K_C$ and it is known beforehand or can be estimated.
Given a coarsely labeled dataset $\mathcal{D}$, our goal is to discover a set of fine-grained classes $\mathcal{Y}_F$.
Thus, we want to recover fine-grained labeling $\tau_F: \mathcal{X} \rightarrow \mathcal{Y}_F$  by using only supervision from the coarsely labeled dataset.





\subsection{Parameterizing the Fine-grained Class Discovery}
\label{subsec:paremeterization}

A key observation in FALCON is that the composition of fine-grained predictions and class relationships produces coarse predictions.
Consequently, we can connect fine predictions and coarse labels using class relations.

We model the fine-grained labeling $\tau_F$ with a probabilistic classifier $f_\theta: \mathcal{X} \rightarrow \Delta^{K_F-1}$ that maps inputs into ($K_F-1$)-dimensional probabilistic simplex $\Delta^{K_F-1}$.
Then, we can recover assignments of samples to fine-grained classes $\mathcal{Y}_F$ by taking the argmax over classifier's fine-grained predictions  $\mathbf{p}_\text{f}$:
\begin{equation}
    \tau_F(\mathbf{x}) = \text{argmax}_i \, \mathbf{p}_\text{f}^i, \quad \text{where} \quad \mathbf{p}_\text{f} = f_\theta(\mathbf{x}).
\end{equation}
Here, $\theta \in \mathbb{R}^d$ parameterizes the fine-grained classifier and  $\mathbf{p}_\text{f}$ is a point on $\Delta^{K_F-1}$. 

We obtain coarse predictions  $\mathbf{p}_\text{c}$ using fine-grained predictions $\mathbf{p}_\text{f}$  and class relations $\mathbf{M}$:
\begin{equation}
    \mathbf{p}_\text{c} = \mathbf{M}^T \mathbf{p}_\text{f},
\end{equation}
where $\mathbf{p}_\text{c}$ is a point on ($K_C-1$)-dimensional probabilistic simplex $\Delta^{K_C-1}$ and $\mathbf{M} \in \{0,1\}^{K_F \times K_C}$ is a binary matrix of fine-coarse class relationships. In particular, element $\mathbf{M}_{ij}$ is $1$ if the $\textit{i}$-th fine-grained class is associated with the $\textit{j}$-th coarse class and $0$ otherwise.
Since every fine-grained class is related to a single coarse class, every matrix row of $\mathbf{M}$ sums to $1$. Thus, $\mathbf{M}$ is the adjacency matrix of an undirected bipartite graph that models relationships between coarse and fine classes.

FALCON simultaneously learns the fine-grained classifier and class relationships by coarse supervision. We use the cross-entropy objective (CE) to leverage coarse supervision and learn parameters $\theta$ and relations $\mathbf{M}$:
\begin{equation}
\label{eq:joint_objective}
    \mathcal{L}_\text{coarse}(\theta, \mathbf{M}|\mathcal{D}) = \frac{1}{|\mathcal{D}|}\sum_{(\mathbf{x}, y_c) \in \mathcal{D}} \text{CE}(\mathbf{M}^Tf_\theta(\mathbf{x}), y_c). 
\end{equation}
Joint optimization of loss (\ref{eq:joint_objective}) w.r.t the discrete class relations $\mathbf{M}$ and continuous classifier parameters $\theta$ is unstable and computationally intensive, as we show in the following sections. 
To avoid these problems,
we extend the objective (\ref{eq:joint_objective}) and conduct alternating optimization of parameters $\theta$ and class relationships $\mathbf{M}$.

The alternating optimization in FALCON is visualized in Figure \ref{fig:task_c2f} and proceeds as follows. 
We first train the fine-grained classifier parameterized by $\theta$, given class relationships $\mathbf{M}$.
Next, we infer class relationships $\mathbf{M}$, given fine-grained predictions of the classifier and coarse labels.
The procedure is repeated for a predefined number of epochs.
The following sections describe the two steps of FALCON's optimization procedure: \textit{(i)} training the fine-grained classifier, and \textit{(ii)} inferring class relationships.
The technical details of the optimization are in Appendix \ref{appenidx:singlesource_alg}.


\subsection{Training Fine-grained Classifier}
\label{subsec:cls_train}

With the class relationships $\mathbf{M}$ fixed, $\mathcal{L}_\text{coarse}(\theta, \mathbf{M}|\mathcal{D})$ becomes $\mathcal{L}_\text{coarse}(\theta|\mathbf{M}, \mathcal{D})$.
Training the fine-grained classifier by solely optimizing (\ref{eq:joint_objective}) with coarse labels is unable to separate fine-grained classes within a coarse class (see Appendix \ref{appendix:coarse_separation} for the detailed analysis).
To overcome this issue, in FALCON we introduce additional objectives that encourage local consistency and confidence of fine-grained predictions, yielding better separation of fine-grained classes within a coarse class.


\textbf{Consistent and confident fine-grained predictions.} 
We enforce consistency in local fine-grained predictions by considering the nearest neighbors of a given input.
We encourage consistent predictions by maximizing the dot product between the predictions for the input sample and the predictions for the neighbouring samples.
The corresponding loss $\mathcal{L}_\text{NN}$ is a log-geometric mean of the dot products \cite{gansbeke20eccv}:
\begin{equation}
\label{eq:nn_loss}
    \mathcal{L}_\text{NN}(\theta|\mathcal{D}) = \frac{-1}{N L}\sum_{(\mathbf{x},y_c) \in  \mathcal{D}} \sum_{\hat{\mathbf{x}} \in  \mathcal{N}(\mathbf{x}, y_c)}  \ln (f_{\theta_\text{EMA}}(\hat{\mathbf{x}})^T f_\theta(\mathbf{x})),
\end{equation}
where $\mathcal{N}(\mathbf{x}, y_c)$ denotes the set of nearest neighbours of a given sample $\mathbf{x}$ within the same coarse class $y_c$, $\hat{\mathbf{x}}$ is an element from $\mathcal{N}(\mathbf{x}, y_c)$, and $L = |\mathcal{N}(\mathbf{x}, y_c)|$.
Parameters $\theta_\text{EMA}$ are computed as an exponential moving average of $\theta$ over iterations \cite{tarvainen17neurips}: 
\begin{equation}
\theta_\text{EMA}^t = \gamma \theta_\text{EMA}^{t-1} + (1-\gamma) \theta^t,
\end{equation}
where $\gamma$ is a hyperparameter and $t$ represents training iterations.
Different from \cite{gansbeke20eccv}, we retrieve the nearest neighbors from the same coarse class and use EMA parameters. 

The loss $\mathcal{L}_\text{NN}$ ensures consistent fine-grained predictions across the neighbouring samples.
However, consistent predictions can also be ambiguous, which prevents the formation of adequate fine-grained classes.
Hence, we encourage a more confident assignment of samples into fine-grained classes by minimizing the cross entropy between the fine-grained predictions and the target distribution $q$:
\begin{equation}
\label{eq:conf_loss}
    \mathcal{L}_\text{conf}(\theta|\mathbf{M}, \mathcal{D}) =  \frac{1}{N}\sum_{(\mathbf{x}, y_c) \, \in \, \mathcal{D}}  \text{CE}(q_{\theta_\text{EMA}}(\mathbf{x}, y_c), f_\theta(\mathbf{x})).
\end{equation}
The fine-grained target distribution $q$ utilizes information from the coarse label $y_c$ to sharpen the distribution over fine-grained classes.
We define the target distribution $q$ using class relations $\mathbf{M}$ and parameters $\theta_\text{EMA}$ as follows:
\begin{equation}
\label{eq:q_target}
    q_{\theta_\text{EMA}}(\mathbf{x},y_c) := \begin{cases}
    \frac{\exp(\mathbf{s}^{y_\text{f}} / T)}{Z}, & \text{if } \mathbf{M}_{y_\text{f},y_c} = 1\\
    0,              & \text{otherwise},
\end{cases}
\end{equation}
where $T$ is a scalar temperature hyperparameter and $\mathbf{s}$ denotes the logits of the fine-grained classifier.
The scalar $Z$ is a normalization constant and is defined as $Z=\sum_{i=1}^{K_F} \mathbf{M}_{i, y_c} \exp( \mathbf{s}^i / T )$. 





The introduced target distribution $q$ and the fine-grained predictions for nearest neighbours can be viewed as a form of pseudolabels, as visualized in Figure \ref{fig:task_c2f} (left). 

We join the loss $\mathcal{L}_\text{NN}$ (\ref{eq:nn_loss}) and the loss $\mathcal{L}_\text{conf}$ (\ref{eq:conf_loss}) into a joint loss $\mathcal{L}_\text{fine}$ that operates over the fine-grained predictions:
\begin{equation}
\label{eq:cons}
    \mathcal{L}_\text{fine}(\theta|\mathbf{M},\mathcal{D}) = \mathcal{L}_\text{NN}(\theta|\mathcal{D}) + \mathcal{L}_\text{conf}(\theta|\mathbf{M},\mathcal{D})
\end{equation}


\textbf{Regularization.}
To avoid degenerate solutions, we further stabilize the training by introducing the entropy maximization loss $\mathcal{L}_\text{reg}$ (\ref{eq:reg_ent}), which is commonly used in clustering-related tasks \cite{gansbeke20eccv,cao22iclr}.
\begin{equation}
\label{eq:reg_ent}
    \mathcal{L}_\text{reg}(\theta|\mathcal{D}) = \ln K_F + \sum_{i=1}^{K_F} \overline{\mathbf{p}}_\text{f}^i \ln \overline{\mathbf{p}}_\text{f}^i, \,\, \overline{\mathbf{p}}_\text{f} = \frac{1}{N} \sum_{\mathbf{x} \in \mathcal{D}} f_\theta(\mathbf{x}).
\end{equation}
The loss $\mathcal{L}_\text{reg}$ helps to avoid degenerate solutions that assign all samples to the same fine class.
 
\textbf{Total loss of the fine-grained classifier.} Putting it all together, FALCON optimizes the following objective for training the fine-grained classifier:
\begin{equation}
\label{eq:final_cls}
    \underset{\theta \, \in \, \mathbb{R}^d}{\text{min}} \left\{ \mathcal{L}(\theta|\mathbf{M}, \mathcal{D}) = \lambda_1 \mathcal{L}_\text{coarse} + \lambda_2 \mathcal{L}_\text{fine} + \lambda_3 \mathcal{L}_\text{reg} \right\},
\end{equation}
where $\lambda_1, \lambda_2$ and $\lambda_3$ are modulation hyperparameters. 
The importance of each loss component is ablated in Section \ref{subsec:ablations}.
Using predictions of the fine-grained classifier, FALCON next learns relationships 
between fine and coarse classes.

\subsection{Inferring Class Relationships}
\label{subsec:class_relationships}
Given the fine-grained classifier $f_\theta$, optimizing (\ref{eq:joint_objective}) involves solving discrete optimization over all possible class relations
to find the optimal $\mathbf{M}$. 
The main difficulties come from the fact that the objective (\ref{eq:joint_objective}) is both a nonlinear function of $\mathbf{M}$ and hard to evaluate due to the large dataset size $N$ ($K_C < K_F \ll N$). Yet, discrete optimization solvers require many evaluations of the objective function and are only effective for specific problem classes such as linear objective functions. To overcome the aforementioned issues, FALCON resorts to the approximation of the objective (\ref{eq:joint_objective}), leading to the efficient inference of the class relationships.


\textbf{Approximated coarse-grained supervision.}
We begin by fixing parameters $\theta$ of the fine-grained classifier and rewriting loss over coarse labels $\mathcal{L}_\text{coarse}$ in the matrix form:
\begin{equation}
\label{eq:cls_matrix_form}
    \mathcal{L}_\text{coarse}(\mathbf{M}|\theta, \mathcal{D}) = - \frac{1}{N} \text{tr}(\mathbf{Y}_{oh}^T \ln(\mathbf{P}\mathbf{M})),
\end{equation}
where $\mathbf{Y}_{oh} \in \{0, 1\}^{N\times K_C}$ is a matrix that represents coarse labels as one-hot vectors and $\mathbf{P} \in [0, 1]^{N\times K_F}$ is a matrix that gathers fine-grained predictions into rows.
The logarithm is applied elementwise while $\text{tr}(\cdot)$ is the trace operator.

To overcome the discussed challenges, we approximate the loss $\mathcal{L}_\text{coarse}$ using Taylor expansion and reformulate it in a computationally efficient way:
\begin{equation}
\label{eq:linear_coarse_cls}
    \mathcal{L}_\text{coarse}^\text{lin}(\mathbf{M}|\theta, D) = - \frac{1}{N} \text{tr}(\mathbf{Y}_{oh}^T\mathbf{P} \mathbf{M}).
\end{equation}
Details of the derivation are provided in Appendix \ref{appendix:derivation_of_lin_cls}.
The cost matrix $\mathbf{C} = \mathbf{Y}_{oh}^T \mathbf{P} \in \mathbb{R}^{K_C \times K_F}_+$
effectively encodes the strength of connections between coarse and fine classes.
Each cost matrix element 
$\mathbf{C}_{ij}$
is proportional to the number of samples from coarse class $j$ assigned to the fine class $i$.
Thus, the optimal solution of (\ref{eq:linear_coarse_cls})
 preserves only the strongest connections between coarse and fine classes.
Note that the objective (\ref{eq:linear_coarse_cls}) can be evaluated more efficiently than (\ref{eq:cls_matrix_form}) since the matrix $\mathbf{Y}_{oh}^T\mathbf{P}$ can be precomputed.

\textbf{Regularization.}
Computing the optimal solution for the objective $\mathcal{L}_\text{coarse}^\text{lin}$ may lead to severely imbalanced assignments of fine classes across coarse classes.
Hence, we introduce an additional regularization term that penalizes the deviation from the balanced assignment of fine-grained classes across coarse classes:
\begin{equation}
\label{eq:M_bal}
    \mathcal{L}_\text{bal}(\mathbf{M}) = \frac{1}{K_C} \text{tr}(\mathbf{M}^T\boldsymbol{1}_{K_F}\boldsymbol{1}_{K_F}^T\mathbf{M}) -  \frac{K_F^2}{K_C^2},
\end{equation}
where  $\boldsymbol{1}_{K_F}$ denotes $K_F$-dimensional column vector of ones.
Thus, $\mathbf{M}^T\boldsymbol{1}_{K_F}$ is a $K_C$-dimensional vector whose values correspond to the number of fine classes associated with every coarse class.
The constant $K_F^2/K_C^2$ corrects the loss so that it yields zero in the case of the balanced assignment.

\textbf{Total loss for inferring class relationships.}  FALCON recovers the relations between fine and coarse classes $\mathbf{M}$ by solving the following optimization problem: 
\begin{equation}
\label{eq:objective_M}
     \underset{\mathbf{M} \, \in \, \mathcal{M}}{\text{min}} \left\{  \mathcal{L}(\mathbf{M}|\theta, \mathcal{D}) =  \mathcal{L}_\text{coarse}^\text{lin}(\mathbf{M}|\theta, D) +  \lambda_M  \mathcal{L}_\text{bal}(\mathbf{M}) \right\},
\end{equation}
where $\lambda_M$ is a hyperparameter that controls the influence of $\mathcal{L}_\text{bal}$.
The set $\mathcal{M}$ contains all possible class relations:
\begin{align}
    \mathcal{M} = \{& \mathbf{M} \in \{0, 1\}^{K_F \times K_C} \, |\, \nonumber \\ & \mathbf{M}\boldsymbol{1}_{K_C} = \boldsymbol{1}_{K_F}, \mathbf{M}^T\boldsymbol{1}_{K_F} \geq \boldsymbol{1}_{K_C} \}.
\end{align}

The optimization problem (\ref{eq:objective_M}) is essentially an integer quadratic program with linear constraints (as discussed in Appendix \ref{appendix:qc_cp}). 
The problem involves the optimization of only $K_F\cdot K_C$ binary variables. 
Thus, we can swiftly compute the solution of (\ref{eq:objective_M}) using modern hardware, even though the resulting problem is inherently NP-hard \cite{fotakis21colt}. 
Our experiments suggest that FALCON can be applied to real-world datasets with hundreds of fine-grained classes.

\subsection{Training on Multiple Datasets}
\label{subsec:multi_datasets}

Fine-grained classes can be grouped into coarse classes in different ways.
For example, one can group animals according to diet (carnivores vs omnivores), size (small vs large), or biological taxonomy (Canis lupus vs Canis familiaris).
Thus, datasets often have different labels despite aggregating the instances of the same fine classes \cite{bevandic24ijcv}. FALCON is seamlessly applicable to training on multiple datasets with different coarse labels.

Specifically, let $\mathcal{D}_l = \{(\mathbf{x}^i, y_c^i)\}_{i=1}^{N_l}$ be a dataset where $\mathbf{x}^i \in \mathcal{X}$, $y_c^i \in \mathcal{Y}_C^l$, and $\mathcal{Y}_C^l$ is  dataset-specific set of coarse classes.
We assume that samples from every dataset $\mathcal{D}_l$ can be associated with fine classes from a shared set of fine classes $\mathcal{Y}_F$.
We aggregate samples from $D$ datasets into a combined dataset $\mathcal{D}_\text{all}$:
\begin{equation}
    \mathcal{D}_\text{all} = \cup_{l=1}^D \{ (\mathbf{x}, y, l) \, | \, (\mathbf{x}, y) \in \mathcal{D}_l \}.
\end{equation}
Every datapoint in $ \mathcal{D}_\text{all}$ is a triplet of input, coarse label, and index of the dataset from which the sample originates.

We extend the parametrization (\ref{eq:joint_objective})
by modeling $D$ dataset-specific mappings $\mathbf{M}_l$:
\begin{multline}
    \mathcal{L}_\text{coarse}(\theta, \mathbf{M}_1, \dots, \mathbf{M}_D | \mathcal{D}_\text{all}) =\\ \frac{1}{|\mathcal{D}_\text{all}|}\sum_{(\mathbf{x}, y_c, l) \in \mathcal{D}_\text{all}} \text{CE}(\mathbf{M}_l^Tf_\theta(\mathbf{x}), y_c) .
\end{multline}
Thus, integrating multiple datasets into the FALCON framework only requires inferring $D$ dataset-specific class relations $M_l$.
As in the case of a single dataset, FALCON infers dataset-specific class relations by solving (\ref{eq:objective_M}).
All $D$ discrete optimization problems are mutually independent and can be solved in parallel.

The fine-grained classifier optimizes the same losses described in Section \ref{subsec:cls_train}.
Still, dataset-specific coarse supervision results in different gradient updates for the same set of fine-grained predictions, resulting in better fine-grained performance, as we show in the experiments.
Technical details for the multi-dataset training procedure are in Appendix \ref{appendix:multisource_alg}.

\section{Experimental Setup}
\label{sec:experiments}


\subsection{Datasets \& Metrics}

\textbf{Datasets.} We evaluate FALCON on eight image classification datasets including Living17, Nonliving26, Entity30, Entity13, tieredImageNet \cite{ren18iclr}, CIFAR100 \cite{krizhevsky09cifar}, CIFAR-SI, and CIFAR68 datasets. 
Datasets Living17, Nonliving26, Entity30 and Entity13 are from the BREEDS benchmark \cite{santurkar21iclr}. 
For the tieredImageNet dataset\cite{ren18iclr}, we joined training, validation and test taxonomies into a single dataset with $608$ fine classes assigned across $34$ coarse classes. 
For the CIFAR100 dataset, we used the original labels with $20$ coarse and $100$ fine classes. 
The original CIFAR100 dataset has an equal number of fine classes associated with every coarse class and an equal number of samples in every fine class.
Hence, we additionally introduce two unbalanced versions of the CIFAR100 dataset that we name CIFAR68 and CIFAR-SI datasets.
In the case of CIFAR68 dataset, we remove $32$ fine classes from the original dataset to disbalance the number of fine classes in coarse classes.
In the case of CIFAR-SI dataset, we remove up to $70\%$ of samples from every fine class, which effectively disbalances sample distribution.
We evaluate performance on the image datasets in the inductive setting on the predefined train/test splits.

In addition, to show that FALCON is widely applicable we consider a single-cell RNA-seq dataset from the biology domain. 
We evaluate FALCON on the PBMC dataset gathered from blood samples of COVID-19 patients \cite{lindeboom23medrxiv}.
The task is to classify cells into fine-grained cell subtypes given coarse-grained cell types. 
We evaluate the method on the ground-truth cell subtypes that correspond to fine-grained labels. 
The PBMC dataset is highly imbalanced (Gini coefficient greater than 0.5). We evaluate performance on the single-cell data in the transductive setting.

The overview of all considered datasets is shown in Table \ref{tab:dataset_info}. 
Abbreviation L17 stands for Living17, N26 for Nonliving26, E30 for Entity30, E13 for Entity13, C100 for CIFAR100, C68 for CIFAR68, CSI for CIFAR Sample Imbalanced, tIN for tieredImageNet, and PB for PBMC.
More details about the datasets are provided in Appendix \ref{appendix:dataset_details}.

\begin{table}[ht]
\vskip -0.1in
\setlength{\tabcolsep}{2pt}
\caption{Overview of nine evaluation datasets. 
The four rows correspond to the dataset abbreviation, the number of coarse classes, the number of fine classes, and the input resolution.
}
\label{tab:dataset_info}
\begin{center}
\begin{small}
\begin{sc}
\begin{tabular}{lccccccccc}
\toprule
Dataset & L17 & N26 & E30 & E13 & C100 & C68 & CSI & tIN  & PB \\
\midrule
\# Coarse & 17 & 26 & 30 & 13 & 20 & 20 & 20 &  34  & 27 \\
\# Fine & 68 & 104 & 240 & 260 & 100 & 68 & 100 & 608  & 83 \\
Res. & $224^2$ & $224^2$ & $224^2$ & $224^2$ & $32^2$ & $32^2$ & $32^2$ & $224^2$ & 2k \\
\bottomrule
\end{tabular}
\end{sc}
\end{small}
\end{center}
\vskip -0.1in
\end{table}

\textbf{Metrics.} We train FALCON and baselines without access to fine-grained ground-truth labels. 
Thus, we report fine-grained clustering accuracy \citep{gansbeke20eccv,vaze22cvpr} as an evaluation metric: 
\begin{equation}
    \text{Acc} = \underset{p \in \mathcal{P}(\mathcal{Y}_\text{f})}{\max} \frac{1}{|D|} \sum_{i=1}^{|D|} \, \llbracket y_\text{f}^i = p(\hat{y}^i_\text{f}) \rrbracket.    
\end{equation}

Here, $\mathcal{P}(\mathcal{Y}_\text{f})$ is a set of all permutations of fine class labels.
In practice, the metric can be efficiently computed using Hungarian algorithm \citep{kuhn55nrl}.
Additionally, we report adjusted rand index (ARI).
Since FALCON also learns class relations, we report the difference between learned label relations and the ground-truth graph using the graph edit distance (GED) \cite{fischer17prl}.
The graph edit distance counts the number of nodes and edges that have to be added or removed in order to match the target graph.

\subsection{Baselines}
\label{subsec:baselines}

Since there is no method specifically design for the setting of fine-grained class discovery with coarse supervision, we compare FALCON with methods that could be applied in this setting, including clustering and cross-granularity few-shot methods that we adapt for fine-grained class discovery.

SCAN \cite{gansbeke20eccv} is a deep clustering method which we directly apply for fine-grained class discovery by clustering the data. However, SCAN cannot utilize information about coarse classes during the training. Thus, we additionally adapt SCAN by enforcing consistent predictions across neighbors retrieved within the same coarse class. This adaptation enables SCAN to utilize coarse supervision. We call this baseline SCAN-C.

We further include cross-granularity few-shot learning methods as baselines. ANCOR \cite{bukchin21cvpr} is a cross-granularity few-shot method which learns fine-grained representation space. Thus, we run K-Means over the extracted features to recover the fine-grained predictions. We use the same approach to adapt SNCA \cite{wu18eccv}. SCGM \cite{ni22iclr} is a few-shot method which can be directly applied to fine-grained class discovery since it provides fine-grained predictions.

We also include GEORGE \cite{sohoni20neurips} which conducts distributionally robust optimization of the coarse classification objective.
GEORGE only learns fine-grained representation space so we again run the K-Means algorithm to recover the fine-grained predictions.


Finally, we can determine the upper bound of the performance by empirical risk minimization (ERM) \cite{vapnik98book} over fine-grained labels.

\subsection{Implementation Details}

We use ResNet18 \cite{he16cvpr} as the backbone for small images from the CIFAR dataset and ResNet50 as the backbone for large images from the remaining five image datasets.
We initialize all methods (\textit{i.e.}, FALCON and all baselines) with self-supervised pretraining method MoCoV3 \cite{chen21iccv}.
We update all parameters of the model during the training.
We pair weak augmentations of the input with $\theta_\text{EMA}$ and strong augmentations with $\theta$, as in \cite{chen21iccv}.
We retrieve nearest neighbours using the distances between self-supervised feature representations.
Hyperparameter search was conducted on the CIFAR100 dataset using the TPE algorithm implemented in Optuna  \cite{akiba19kdd}.
We solve the discrete optimization problem using Gurobi \cite{gurobi23manual}.

In the case of single-cell data, we use randomly initialized MLP with 4 linear layers and ReLU activations.
We retrieve the nearest neighbours by computing the distance over the top 2k highly variable genes.
More implementation details are in the Appendix \ref{appendix:implementation_details}. 
Our code is publicly available\footnote{\url{https://github.com/mlbio-epfl/falcon}}.

\section{Experimental Evaluation}


\begin{table*}[ht]
\vskip -0.1in
\setlength{\tabcolsep}{3pt}
\caption{Fine-grained accuracy (Acc) and adjusted rand index (ARI) on eight image datasets and a single-cell RNA-seq dataset. Results are averaged over three runs. Transductive evaluation is denoted with $\dagger$.}
\label{tab:fine-grained-cls}
\begin{center}
\begin{small}
\begin{sc}
\begin{tabular}{lcc|cc|cc|cc|cc|cc|cc|cc|cc}
\toprule
\multirow{2}{*}{Method} & \multicolumn{2}{c|}{Living17} & \multicolumn{2}{c|}{NLiving26} & \multicolumn{2}{c|}{Entity30} & \multicolumn{2}{c|}{Entity13} & \multicolumn{2}{c|}{CIFAR100} & \multicolumn{2}{c|}{CIFAR68} & \multicolumn{2}{c|}{CIFAR-SI} & \multicolumn{2}{c|}{tieredIN}  & \multicolumn{2}{c}{$\text{PBMC}^\dagger$} \\
 & Acc & ARI & Acc & ARI & Acc & ARI & Acc & ARI & Acc & ARI & Acc & ARI & Acc & ARI & Acc & ARI & Acc & ARI \\
 \midrule
Up. Bound & 86.3 & 76.1 & 84.6 & 73.8 & 85.6 & 74.5 & 85.9 & 75.4 & 74.5 & 57.0 & 78.8 & 62.9 & 73.0 & 56.6 & 79.1 &  65.1 & 88.9 & 86.4\\[0.5em]
SCAN & 61.9 & 50.1 & 54.3 & 39.7 & 51.1 & 38.4 & 50.8 & 37.5 & 47.1 & 34.4 & 51.4 & 39.8 & 47.4 & 35.8 & 43.6 & 28.9  & 20.7 & 13.5\\
ANCOR & 27.7 & 36.1 & 27.9 & 34.7 & 17.0 & 20.2 & 8.4 & 8.5 & 23.4 & 26.6 & 30.1 & 33.7 & 28.7 & 25.9 & 47.8 & 34.1  & 44.9 & 37.7 \\
SNCA & 39.2 & 30.9 & 43.6 & 31.1 & 36.1 & 23.4 & 35.1 & 20.9 & 42.9 & 18.9 & 47.6 & 23.3 & 41.3 & 21.6 & 22.3 & 11.0  & 29.5 & 20.2 \\
GEORGE &  62.8 & 53.2 & 58.8 & 47.2 & 50.1 & 35.6 & 49.6 & 35.7 & 51.9 & 36.0 & 59.6 & 42.8 & 51.2 & 36.7 & 43.0 & 29.1 & 37.7 & 32.4 \\
SCGM & 62.3 & 49.3 & 56.4 & 42.0 & 56.0 & 41.4 & 54.8 & 40.8 & 47.9 & 32.2 & 49.6 & 34.7 & 46.3 & 31.8 &  46.6 &  32.0  & 22.7 & 13.8\\
SCAN-C & 67.1 & 54.7 & 60.4 & 45.8 & 60.6 & 46.2 & 57.7 & 43.7 & 48.7 & 36.1 & 54.3 & 41.9 & 48.6 &38.0 & 48.2 & 33.2 & 20.3 & 13.2 \\[0.5em]
FALCON & \textbf{71.8} & \textbf{60.3} & \textbf{65.7} & \textbf{55.5} & \textbf{65.1} & \textbf{53.3} & \textbf{63.6} & \textbf{51.9} & \textbf{59.6} & \textbf{42.5} & \textbf{60.4} & \textbf{47.0} & \textbf{55.6} & \textbf{39.1} & \textbf{53.4} & \textbf{41.6} & \textbf{75.8} & \textbf{74.0} \\
\bottomrule
\end{tabular}
\end{sc}
\end{small}
\end{center}
\vskip -0.2in
\end{table*}

\subsection{Quantitative Results}
\label{subsec:fg_classification}

\textbf{Fine-grained class discovery.} 
We compare FALCON to alternative baselines on eight image classification datasets and a single-cell dataset. The results in Table \ref{tab:fine-grained-cls} show that FALCON outperforms baselines by a large margin on both image and single-cell data. For example, on the BREEDS benchmark of four datasets (Living17, Nonliving26, Entity30, Entity13) FALCON achieves an average improvement of $9\%$ in terms of accuracy and $16\%$ in terms of ARI over the best baselines. On the tieredImageNet dataset with $608$ fine-grained classes grouped into $34$ coarse classes, FALCON outperforms the best baseline ANCOR by $12\%$ and $22\%$ in terms of accuracy and ARI, respectively. Moreover, improvements of FALCON can be observed on both balanced (BREEDS benchmark and CIFAR100), as well as unbalanced datasets (CIFAR68, CIFAR-SI, tieredImageNet and single-cell PBMC).

\textbf{Evaluation of learned class relationships.}
FALCON infers the mapping between fine-grained and coarse-grained classes. We next evaluate how well the inferred coarse-fine class relationships agree with the ground truth relationships using the graph edit distance (GED).
The graph edit distance of zero indicates that two graphs are the same and the ground truth relationships are perfectly inferred.
We compare FALCON with SCGM which learns class relations implicitly through a data generation process. 
The results shown in Table \ref{tab:abl_num_fine_grained} demonstrate that FALCON
can perfectly match the ground truth relations for all balanced datasets which is not the case for the SCGM.
For the imbalanced datasets, not all class relationships are correctly recovered but FALCON still substantially outperforms SCGM on all datasets.
The discrepancy between the learned and the ground-truth relations comes from the fact that (\ref{eq:objective_M}) finds the optimal solution given a classifier $f_\theta$.
Thus, the obtained solution may differ from the true optimal solution.
This discrepancy could be mitigated by employing a stronger classifier \cite{radford21icml}.
\begin{table}[htb]
\setlength{\tabcolsep}{3pt}
\caption{Graph edit distance (GED) between the learnt class relations and the true class relations averaged over three runs.}
\label{tab:abl_num_fine_grained}
\setlength{\tabcolsep}{2pt}
\begin{center}
\begin{small}
\begin{sc}
\vskip -0.15in
\resizebox{\columnwidth}{!}{
\begin{tabular}{lccccccccc}
\toprule
GED$\downarrow$ & L17 & N26 & E30 & E13 & C100 & C68 & CSI & tIN & PM\\
\midrule
SCGM & 30.7 & 74.7 & 98.7 & 45.3 & 61.3 & 57.0 & 61.3 & 132.0 & 88.7\\
FALCON & \textbf{0.0} & \textbf{0.0} & \textbf{0.0} & \textbf{0.0} & \textbf{0.0} & \textbf{40.3} & \textbf{0.0} & \textbf{110.7} & \textbf{73.3} \\
\bottomrule
\end{tabular}
}
\end{sc}
\end{small}
\end{center}
\vskip -0.2in
\end{table}

\textbf{Training on multiple datasets.}
FALCON can learn from multiple datasets labeled with different coarse-level classes (Section \ref{subsec:multi_datasets}). We evaluate the effectiveness of FALCON on the CIFAR100 dataset. The CIFAR100 dataset has a default grouping of $100$ fine classes into $20$ coarse classes which we denote with T1.
We construct a meaningful alternative grouping of fine classes into coarse classes, which we denote as T2.
For example, the default grouping T1 arbitrarily divides $10$ fine-grained vehicle classes into two coarse classes named \textit{Vehicles1} and \textit{Vehicles2}.
On the contrary, our alternative taxonomy groups fine-grained vehicle classes into  \textit{Personal Vehicles} and \textit{Transit Vehicles} (see Appendix \ref{appendix:cifar100_taxonomies} for the full list of coarse classes in each taxonomy).
We split the training set into two halves and label the first half according to T1 and the second half according to T2.
Thus, the resulting splits, denoted with $\mathcal{D}_1$ and $\mathcal{D}_2$, correspond to two datasets with the same underlying set of fine-grained classes and different coarse classes.
We keep the CIFAR100 test set unmodified to track the generalization performance for different training configurations.

Table\ \ref{tab:multiple_sources1} shows the results of using FALCON to simultaneously learn from two coarsely labeled datasets with different labeling policies.
The top two rows show fine-grained accuracy and ARI after training FALCON on $\mathcal{D}_1$ or $\mathcal{D}_2$.
Compared to FALCON trained on a single dataset, 
we observe $14\%$ relative improvement according to ARI and $10\%$ relative improvement according to clustering accuracy.
These results indicate that FALCON can effectively utilize different labeling policies to improve performance. 

To evaluate the benefits of training from multiple datasets using baseline methods, we train SCAN-C on the same datasets.
While SCAN-C can simultaneously learn from multiple datasets, the gains from different labelings are marginal compared to FALCON (2\% improvement in terms of ARI).
\begin{table}[ht]
\setlength{\tabcolsep}{3pt}
\caption{ FALCON benefits from simultaneous training on multiple datasets with incompatible coarse labels.}
\label{tab:multiple_sources1}
\begin{center}
\begin{small}
\begin{sc}
\begin{tabular}{ccccccc}
\toprule
\multirow{2}{*}{Train DS}& \multirow{2}{*}{Samples} & \multirow{2}{*}{Taxonomy} & \multicolumn{2}{c}{SCAN-C} & \multicolumn{2}{c}{FALCON} \\
 & &  & Acc & ARI & Acc & ARI \\
\midrule
$\mathcal{D}_1$ & N/2 & T1 & 48.7 & 35.9 & 56.0 & 38.6  \\
$\mathcal{D}_2$ & N/2 & T2 & 47.8 & 34.4 & 56.6 & 38.7 \\[.3em]
$\mathcal{D}_\text{all}$ & N & T1\&T2 & 49.6 & 36.7 & \textbf{61.5} & \textbf{43.9} \\
\bottomrule
\end{tabular}
\end{sc}
\end{small}
\end{center}
\vskip -0.2in
\end{table}

Aggregating multiple training datasets implicitly increases the number of training samples and thus improves the generalization.
Hence, we analyze the influence of different labeling policies in isolation by
repeating the experiment with an equal number of samples in each training dataset from Table \ref{tab:multiple_sources1}.
The results, summarized in Appendix \ref{appendix:multisource_additional_results}, confirm that FALCON can effectively utilize the different labeling policies.
For example, FALCON trained on taxonomies T1\&T2 improves 7\% in terms of accuracy over the FALCON trained only on T2.
Contrary, SCAN-C trained on T1\&T2 improves only 1\% over the SCAN-C trained only on T2.
Altogether, these results indicate that FALCON can effectively learn from multiple datasets labeled with different labeling policies.

\subsection{Qualitative Results}

\begin{figure}[ht]
\vskip -0.1in
\begin{center}
\centerline{\includegraphics[width=\columnwidth]{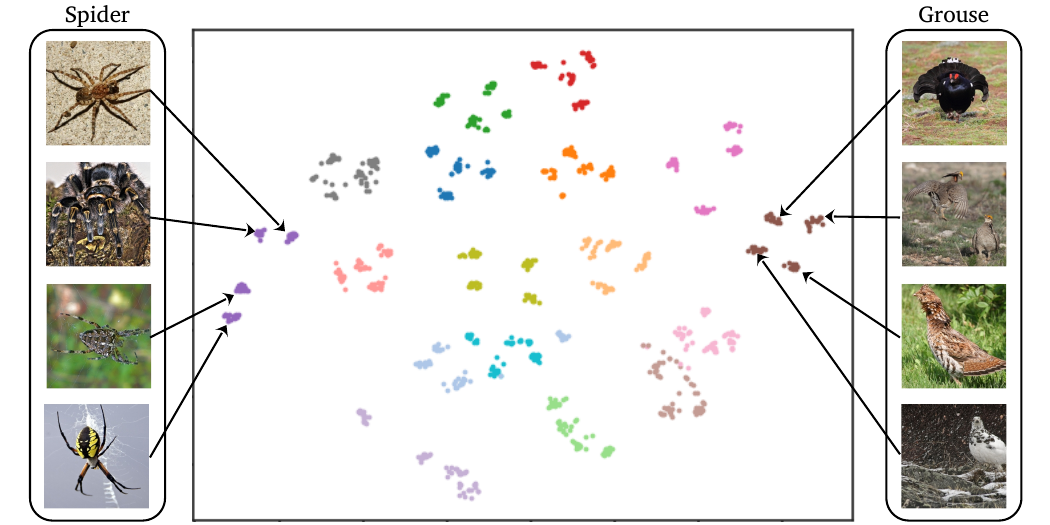}}
\vskip -0.1in
\caption{
The t-SNE plot of Living17 test samples in the embedding space learned by FALCON. Coarse-grained classes used to supervise the model are shown in different colors. The images at the left and right side show representative examples of inferred fine-grained classes for  coarse classes \textit{Spider} and \textit{Grouse}.
}
\label{fig:tsne}
\end{center}
\vskip -0.25in
\end{figure}

We next visually inspect the quality of the embedding space learnt by FALCON. Figure \ref{fig:tsne} shows two-dimensional t-SNE plot \cite{hinton02neurips} of the embedding space learnt by FALCON on the Living17 dataset. 
The results show that samples assigned to the same coarse-grained classes are separated into multiple fine-grained classes. 

To validate that these fine classes are correct and represent different subspecies of animals, we look into representative examples from every fine-grained class and confirm that the examples indeed reflect different subcategories. 
For example, the four fine classes of coarse class \textit{Spider}  correspond to subspecies of spiders including \textit{Barn spider}, \textit{Tarantula}, \textit{Garden spider}, and \textit{Black and gold spider}.
Similarly, the inferred fine-grained classes of \textit{Grouse} correspond to \textit{Black grouse}, \textit{Prairie grouse}, \textit{Ruffed grouse} and \textit{Ptarmigan}.
This indicates that the embedding space learnt by FALCON indeed reflects fine-grained representations.

\begin{figure}[ht]
\begin{center}
\centerline{\includegraphics[width=\columnwidth]{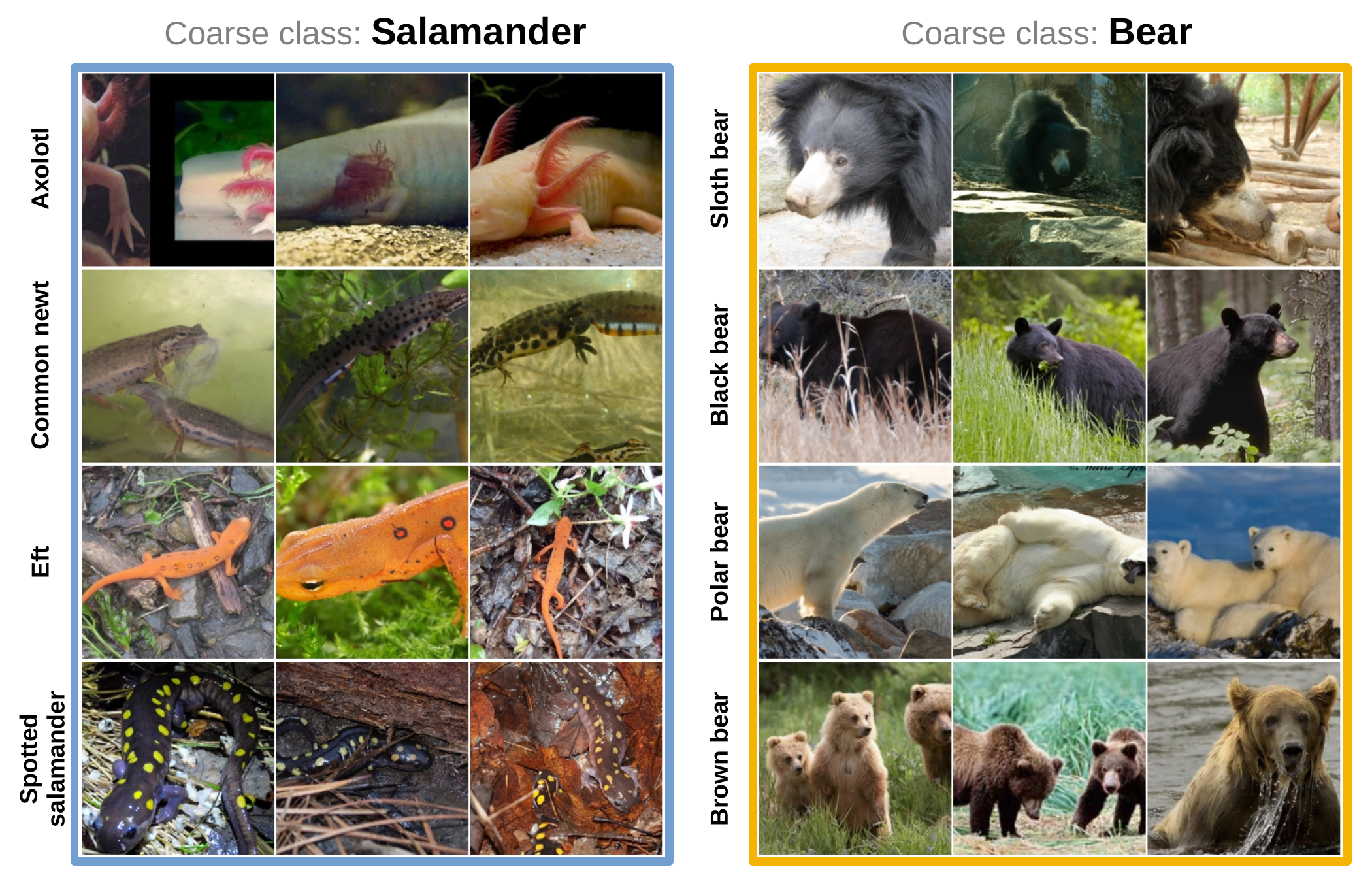}}
\caption{Three most confident predictions for fine-grained classes associated with coarse class \textit{Salamander} and \textit{Bear}.
}
\label{fig:cluster_visual_samples}
\end{center}
\vskip -0.4in
\end{figure}

We next visualize the three most confident predictions for different fine-grained classes. Figure \ref{fig:cluster_visual_samples} shows the three most confident samples for every fine-grained class  associated with coarse classes \textit{Salamander} and \textit{Bear} from the Living17 dataset. 
We validate the recovered fine-grained classes and confirm that they indeed correspond to different salamander subspecies (\textit{Axolotl}, \textit{Common newt}, \textit{Eft},  and \textit{Spotted salamander}) and bear subspecies (\textit{Sloth bear}, \textit{Black bear}, \textit{Polar bear}, and \textit{Brown bear}).
This indicates that FALCON produces meaningful fine-grained classes even when differences between these subclasses are very subtle.
We show more examples in the Appendix \ref{appendix:visual_examples_more}.

\begin{figure}[ht]
\begin{center}
\centerline{\includegraphics[width=\columnwidth]{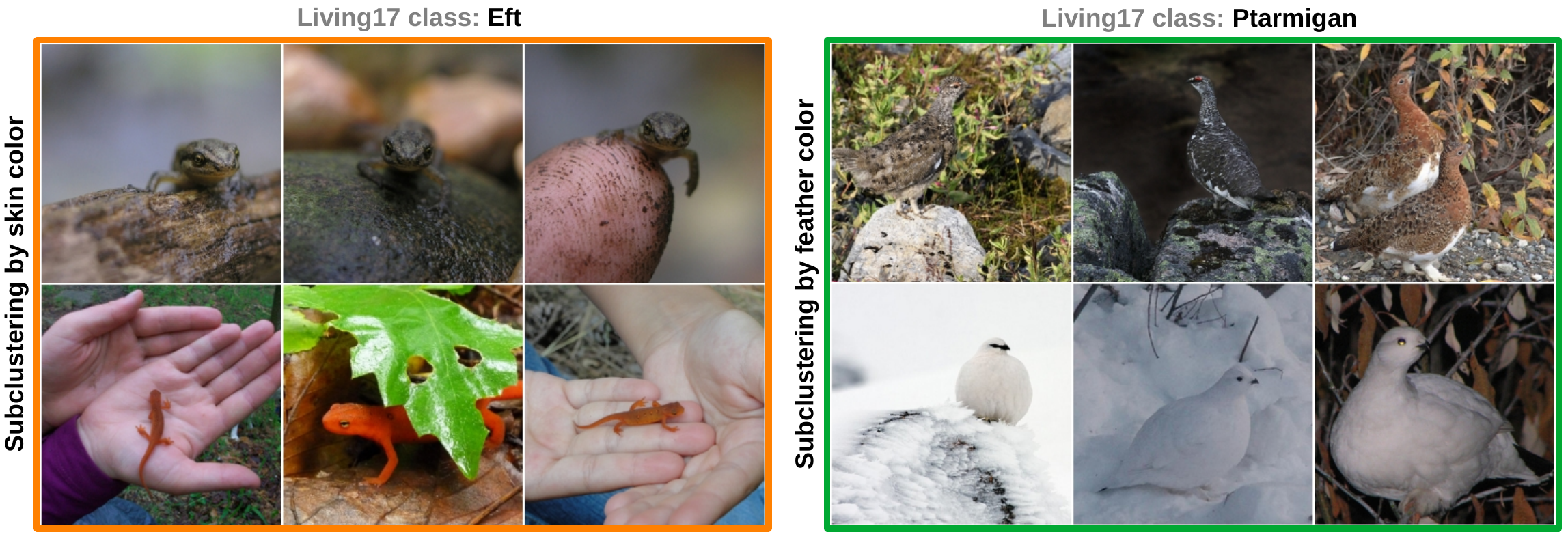}}
\vskip -0.1in
\caption{Subclasses discovered within fine-grained classes \textit{Eft} and \textit{Ptarmigan}.
The discovered subclasses differ according to the skin (\textit{Eft}) or feather (\textit{Ptarmigan}) color.
}
\label{fig:l17_subclasses}
\end{center}
\vskip -0.4in
\end{figure}

FALCON can also discover subclasses within existing fine-grained classes.
To showcase this, we set 68 fine classes of the Living17 dataset as coarse classes and increased the expected number of fine-grained classes.
Figure \ref{fig:l17_subclasses} shows the two subclasses discovered within classes \textit{Eft} and \textit{Ptarmigan}. 
The newly discovered subclasses differ by skin and feather color.
Unfortunately, this evaluation can only be qualitatively verified since the appropriate ground-truth is unavailable.
More visual examples can be found in Appendix \ref{appendix:more_visual_subclasses_l17}.

\subsection{Ablation Studies}
\label{subsec:ablations}

\textbf{Components of the objective.} We next conduct ablation studies on the classifier's objective function (\ref{eq:final_cls}) which consists of coarse supervision $\mathcal{L}_\text{coarse}$, fine-grained consistency and confidence $\mathcal{L}_\text{fine}$, and entropy regularization $\mathcal{L}_\text{reg}$.
To investigate the importance of each part, we modify FALCON by removing each loss component and then measure fine-grained clustering accuracy and ARI.
We show the results on the CIFAR100 dataset in Table \ref{tab:abl_loss_components}.
Removing $\mathcal{L}_\text{fine}$ results in divergent fine-grained predictions for similar samples.
Thus, the samples are arbitrarily grouped into subclasses and we observe poor fine-grained performance.
Removing $\mathcal{L}_\text{reg}$ results in a skewed distribution of samples across fine-grained classes and poor fine-grained performance.
Removing $\mathcal{L}_\text{coarse}$ eliminates coarse-level supervision and prevents joint learning of class relationships.
Thus we also have to remove $\mathcal{L}_\text{conf}$ since it relies on class relations.
Again, we observe a notable performance drop.
This ablation confirms that all three losses contribute to strong fine-grained performance.
\begin{table}[ht]
\vskip -0.2in
\setlength{\tabcolsep}{7pt}
\caption{ Ablation study of the components from the objective function (\ref{eq:final_cls}) on the CIFAR100 dataset.  We report fine-grained performance averaged over three runs by removing different components of the FALCON objective.}
\label{tab:abl_loss_components}
\begin{center}
\begin{small}
\begin{sc}
\begin{tabular}{ccccc}
\toprule
$\mathcal{L}_\text{coarse}$ & $\mathcal{L}_\text{fine}$ & $\mathcal{L}_\text{reg}$ & Acc & ARI \\
\midrule
\cmark & \xmark &  \cmark & 22.9 & 14.6\\
\cmark & \cmark & \xmark & 18.7 & 22.8 \\
\xmark & \cmark & \cmark & 51.0  & 32.0 \\
 \cmark & \cmark & \cmark & \textbf{59.6} & \textbf{42.5} \\
\bottomrule
\end{tabular}
\end{sc}
\end{small}
\end{center}
\vskip -0.2in
\end{table}

\textbf{Estimated number of fine-grained classes.} 
FALCON takes the expected number of fine-grained classes as a hyperparameter. However, in practice, we often do not know the number of classes in
advance. In such a case, we can first estimate the number of novel classes. We estimate the number of classes as proposed in \cite{wang18aaai} and obtain $89$ for the CIFAR100 dataset and $76$ for the CIFAR68 dataset. We then train FALCON with the estimated number of fine-grained classes $K_F$. The results are shown in Table \ref{tab:abl_estimated_kf}.

Remarkably, on the CIFAR100 dataset, FALCON with the estimated number of classes outperforms all other baselines trained with the ground truth number of fine classes. On the CIFAR68 dataset, FALCON outperforms all baselines except GEORGE which attains slightly better accuracy.
FALCON's sensitivity on different values of $K_F$ can be found in Appendix \ref{appendix:sensitivity_on_number_of_fine}. 
\begin{table}[ht]
\vskip -0.2in
\setlength{\tabcolsep}{7pt}
\caption{FALCON performance with the estimated number of fine-grained classes.}
\label{tab:abl_estimated_kf}
\begin{center}
\begin{small}
\begin{sc}
\begin{tabular}{lcccc}
\toprule
\multirow{2}{*}{Method} & \multicolumn{2}{c}{CIFAR100} & \multicolumn{2}{c}{CIFAR68} \\
 & Acc & ARI & Acc & ARI \\
\midrule
GEORGE & 51.9 & 36.0 & 59.6 & 42.8 \\[0.5em]
FALCON & 59.6 & 42.5 & 60.4 & 47.0 \\
FALCON-$\hat{K_F}$ & 54.9 & 39.9 & 59.1 & 45.7 \\
\bottomrule
\end{tabular}
\end{sc}
\end{small}
\end{center}
\vskip -0.2in
\end{table}

\textbf{Additional ablation studies.} We validate the fine-grained performance for different hyperparameter values on CIFAR100.
We validate  $\lambda_M$ from  (\ref{eq:objective_M}), temperature $T$ from (\ref{eq:q_target}), and three loss modulation hyperparameters $\lambda_1, \lambda_2$ and $\lambda_3$ from (\ref{eq:final_cls}).
The results are summarized in Appendix \ref{appendix:hyperparameters_sensitiviy}.
FALCON is robust to different values of 
hyperparameters.
In addition, we evaluate the performance of FALCON with
ground truth class relations in Appendix \ref{appendix:estim_vs_actual}.
The results indicate that estimating class relations does not significantly affect the fine-grained classification performance.
Thus, knowing class relations beforehand is not mandatory for good fine-grained performance.

\section{Conclusion}

We presented FALCON, a method that discovers fine-grained classes within coarsely labeled data.
FALCON can simultaneously learn fine-grained classifier and relationships between the discovered fine and the available coarse classes by coarse supervision.
We combine the coarse classification loss with additional fine-grained consistency and regularization losses to discover meaningful fine-grained classes.
Simultaneously, we infer relations between the discovered fine and the available coarse classes by solving a discrete optimization problem.
Such modular design enables FALCON to simultaneously train on multiple datasets with different coarse labels. 
FALCON consistently outperforms all baselines on large-scale image classification datasets and a single-cell dataset from the biology domain.

\section{Limitations}
\label{sec:limitations}

\textbf{Number of fine-grained classes.} FALCON assumes that the number of fine-grained classes is known a priori, which is a common assumption in clustering and open-world learning problems  \cite{gansbeke20eccv,vaze22cvpr,cao22iclr, gadetsky2023pursuit}.
Still, FALCON can be combined with the existing methods which estimate this quantity \cite{wang18aaai} and deliver competitive results, as show in Section \ref{subsec:ablations}.
Furthermore, FALCON is robust to noisy estimates of $K_F$, as discussed in Appendix \ref{appendix:sensitivity_on_number_of_fine}.

\textbf{Consistently labeled dataset.} 
FALCON assumes that samples within the same dataset are consistently labeled. 
That is, instances of the same fine-grained classes are always labeled as the same coarse class. 
However, FALCON can be simultaneously trained on multiple datasets with inconsistent labeling policies, as shown in Section \ref{subsec:fg_classification}.
Thus, instances of the same fine-grained class can be differently labeled in different datasets.


\textbf{The datasets are not severely imbalanced.}
The introduced entropy regularization term in FALCON encourages uniform distribution or samples to fine-grained classes. Since the influence of regularization is controlled by a hyperparameter, 
FALCON is still applicable even on sample-imbalanced datasets as we show in our experiments.
However, extremely long-tailed class distribution will require replacing entropy regularization with alternative regularization objectives such as KL divergence between empirical label distribution and prior over the target label distribution.



\section*{Acknowledgements}

The authors thank Ramón Viñas Torné and Yulun Jiang for the useful comments on the manuscript, as well as Maciej Wiatrak for suggesting an adequately annotated single-cell dataset. We gratefully acknowledge the support of EPFL and ZEISS. Matej Grcić was supported by the Swiss Federal Government Excellence scholarship for foreign students.

\section*{Impact Statement}
This paper presents work whose goal is to advance the field of Machine Learning. There are many potential societal consequences of our work, non which we feel must be specifically highlighted here.

\bibliography{main}
\bibliographystyle{icml2024}

\newpage
\appendix
\onecolumn



\section{FALCON Training Algorithm}
\label{appenidx:singlesource_alg}
\renewcommand{\thetable}{\Alph{section}\arabic{table}}
\renewcommand\thefigure{\Alph{section}\arabic{figure}} 
\renewcommand\thealgorithm{\Alph{section}\arabic{algorithm}}
\renewcommand{\theHtable}{\Alph{section}\arabic{table}}
\renewcommand\theHfigure{\Alph{section}\arabic{figure}} 
\renewcommand\theHalgorithm{\Alph{section}\arabic{algorithm}}
\setcounter{table}{0}
\setcounter{figure}{0}
\setcounter{algorithm}{0}

Algorithm \ref{alg:fg_cls} shows the training procedure for simultaneous learning of a fine-grained classifier and class relationships.
We initialize feature extractor with self-supervised pretraining \cite{chen21iccv} in the case of images and randomly in the case of single-cell data.
We construct a fine-grained classifier by appending a linear classifier atop the feature extractor.
We initialize $\mathbf{M}$ by solving (\ref{eq:objective_M}) for a randomly initialized cost matrix.
In every epoch, we iteratively update parameters $\theta$ using SGD over the sampled minibatch. 
After every P iterations, we update class relations $\mathbf{M}$ using the current values of $\theta$.
We do not gather predictions for the whole dataset but rather for a large enough subset (\textit{i.e.}~$20\times$ batch size).
The training of $\theta$ then progresses with the new value of $\mathbf{M}$. Steps of the algorithm are outlined in Algorithm 1.

\begin{algorithm}[ht] 
   \caption{FALCON training on a single dataset}
   \label{alg:fg_cls}
\begin{algorithmic}
   \STATE {\bfseries Input:} Fine-grained classifier $f_\theta$, dataset $\mathcal{D}$, number of fine-grained classes $K_F$, hyperparameters $\boldsymbol{\lambda}, T, \lambda_M$, $\Omega$
   \STATE $\theta = \text{selfsup\_initialization}(\mathcal{D})$ 
   \STATE $\mathbf{M}$ = solve (\ref{eq:objective_M}) for random cost matrix
   \FOR{$i=1$ {\bfseries to} max\_epochs}
   \FOR{$j=1$ {\bfseries to} max\_iters}
   \STATE Sample minibatch $(\mathbf{x}, y)$ from $\mathcal{D}$
   \STATE $L_\text{cls}$ =  evaluate loss (\ref{eq:joint_objective})
    \STATE $L_\text{cons}$ =  evaluate loss (\ref{eq:cons})
   \STATE $L_\text{reg}$ =  evaluate loss (\ref{eq:reg_ent})
   \STATE $L = \lambda_1 \cdot L_\text{cls} + \lambda_2 \cdot L_\text{cons} + \lambda_3 \cdot L_\text{reg}$
   \STATE $\theta = \text{SGD}(L, \theta)$
   \IF{j \% $P$ = 0 
   }
   \STATE Gather $\mathbf{Y}_{oh}$ and $\mathbf{P}$ using $f_\theta$ over a subset of $\mathcal{D}$
   \STATE $\mathbf{C} = \mathbf{Y}_{oh}^T\mathbf{P}$
   \STATE $\mathbf{M}$ =  solve (\ref{eq:objective_M}) for cost matrix $\mathbf{C}$
   \ENDIF
   \ENDFOR
   \ENDFOR
\STATE {\bfseries Output:} classifier parameters $\theta$, class relationships $\mathbf{M}$
\end{algorithmic}
\end{algorithm}

\section{Impact of Coarse Supervision to Fine-grained Predictions}
\label{appendix:coarse_separation}
\renewcommand{\thetable}{\Alph{section}\arabic{table}}
\renewcommand\thefigure{\Alph{section}\arabic{figure}} 
\renewcommand\thealgorithm{\Alph{section}\arabic{algorithm}}
\renewcommand{\theHtable}{\Alph{section}\arabic{table}}
\renewcommand\theHfigure{\Alph{section}\arabic{figure}} 
\renewcommand\theHalgorithm{\Alph{section}\arabic{algorithm}}
\setcounter{table}{0}
\setcounter{figure}{0}
\setcounter{algorithm}{0}
Let $\mathbf{p}_\text{f} \in \Delta^{K_F-1}$ 
be the output of the fine-grained classifier $f_\theta$ for a given training example $\mathbf{x}$,  $\mathbf{p}_\text{c} \in \Delta^{K_C-1}$ be the output of coarse-grained classifier $\mathbf{M}^Tf_\theta$, and $j$ be the corresponding ground-truth coarse label.
The loss $\mathcal{L}_\text{coarse}$ (\ref{eq:joint_objective}) for a particular sample can be written as:
\begin{equation}
    \mathcal{L}_\text{coarse}(\theta | \mathbf{M}, \mathbf{x}, y_c=j) = \text{CE}(\mathbf{p}_\text{c}, j), \quad \text{where} \quad \mathbf{p}_\text{c}=\mathbf{M}^T\mathbf{p}_\text{f} \quad \text{and} \quad \mathbf{p}_\text{f}=f_\theta(\mathbf{x}).
\end{equation}
The gradient of $\mathcal{L}_\text{coarse}$ w.r.t $i$-th fine prediction equals to: 
\begin{equation}
    \frac{\partial \mathcal{L}_\text{coarse}}{\partial \mathbf{p}_\text{f}^i} = \frac{\partial \mathcal{L}_\text{coarse}}{\partial \mathbf{p}_\text{c}^j} \frac{ \partial \mathbf{p}_\text{c}^j}{\partial \mathbf{p}_\text{f}^i} = \frac{\partial \mathcal{L}_\text{coarse}}{\partial \mathbf{p}_\text{c}^j} \mathbf{M}_{ij}.
\end{equation}
Since $M$ is a binary matrix, the gradient of loss w.r.t $i$-th fine class boils down to: 
\begin{equation}
    \frac{\partial \mathcal{L}_\text{coarse}}{\partial \mathbf{p}_\text{f}^i} = \begin{cases}
    \frac{\partial \mathcal{L}_\text{coarse}}{\partial \mathbf{p}_\text{c}^j}, & \text{if } \mathbf{M}_{ij} = 1\\
    0,              & \text{otherwise}
    \end{cases}
\end{equation}
Thus, all fine-grained classes associated with coarse label $j$ receive the same gradient.
Consequently, $\mathcal{L}_\text{coarse}$ only separates fine-grained classes that are associated with different coarse classes.
We combine the coarse classification loss with additional fine-grained losses that further encourage the separation of fine classes.

\section{Step-by-step derivation of $\mathcal{L}^\text{lin}_\text{cls}$}
\label{appendix:derivation_of_lin_cls}
\renewcommand{\thetable}{\Alph{section}\arabic{table}}
\renewcommand\thefigure{\Alph{section}\arabic{figure}} 
\renewcommand\thealgorithm{\Alph{section}\arabic{algorithm}}
\renewcommand{\theHtable}{\Alph{section}\arabic{table}}
\renewcommand\theHfigure{\Alph{section}\arabic{figure}} 
\renewcommand\theHalgorithm{\Alph{section}\arabic{algorithm}}
\setcounter{table}{0}
\setcounter{figure}{0}
\setcounter{algorithm}{0}

We start from the cross entropy loss in the matrix form:
\begin{equation}
    \mathcal{L}_\text{coarse}(\mathbf{M}|\theta, D) = - \frac{1}{N} \text{tr}(\mathbf{Y}_{oh}^T \ln(\mathbf{P}\mathbf{M})).
\end{equation}
The logarithm function can be expressed with the Taylor series for $0 <x \leq 2$ as:
\begin{equation}
    \ln x = \sum_{i=1}^\infty \frac{(-1)^{i+1}}{i} (x - 1)^i.
\end{equation}
All row vectors of matrix $\mathbf{P}$ are points on $\Delta^{K_F-1}$ and all entries of $\mathbf{M}$ are either 0 or 1 therefore elements of $\mathbf{P}\mathbf{M}$ are positive and less or equal to 1. 
Consequently, we can approximate the logarithm function with the Taylor series truncated after the first term.
\begin{align}
    \mathcal{L}_\text{coarse}(M|\theta, D) &= - \frac{1}{N} \text{tr}(\mathbf{Y}_{oh}^T \ln(\mathbf{P}\mathbf{M})) \\
    &\approx - \frac{1}{N} \text{tr}(\mathbf{Y}_{oh}^T (\mathbf{P}\mathbf{M} - \boldsymbol{1}_{N \times K_C})) \\
    &= - \frac{1}{N} \text{tr}(\mathbf{Y}_{oh}^T \mathbf{P}\mathbf{M} - \mathbf{Y}_{oh}^T\boldsymbol{1}_{N \times K_C}) \\
    & = - \frac{1}{N} \text{tr}(\mathbf{Y}_{oh}^T \mathbf{P}\mathbf{M}) + \frac{1}{N} \text{tr}(\mathbf{Y}_{oh}^T\boldsymbol{1}_{N \times K_C}) \\
    & = - \frac{1}{N} \text{tr}(\mathbf{Y}_{oh}^T \mathbf{P}\mathbf{M}) + 1 \\
    & \cong - \frac{1}{N} \text{tr}(\mathbf{Y}_{oh}^T \mathbf{P}\mathbf{M})
\end{align}

\section{Class Relationships by Solving Integer  Quadratic Program}
\label{appendix:qc_cp}
\renewcommand{\thetable}{\Alph{section}\arabic{table}}
\renewcommand\thefigure{\Alph{section}\arabic{figure}} 
\renewcommand\thealgorithm{\Alph{section}\arabic{algorithm}}
\renewcommand{\theHtable}{\Alph{section}\arabic{table}}
\renewcommand\theHfigure{\Alph{section}\arabic{figure}} 
\renewcommand\theHalgorithm{\Alph{section}\arabic{algorithm}}
\setcounter{table}{0}
\setcounter{figure}{0}
\setcounter{algorithm}{0}

Integer quadratic programs \cite{boyd14book} are programs of the following form:
\begin{equation}
    \underset{\mathbf{x}}{\text{min}} \quad \mathbf{x}^TA_0\mathbf{x} + \mathbf{b}_0^T\mathbf{x} + c_0 
\end{equation}
subjected to inequality and equality constraints:
\begin{equation}
    \mathbf{b}_i^T\mathbf{x} + c_i \leq 0, \quad i=1,\dots, m
\end{equation}
\begin{equation}
    \mathbf{d}_j\mathbf{x} + e_j = 0, \quad j=1,\dots,n.
\end{equation}
The matrix $A$ is symmetric positive semidefinite, all $\mathbf{b}_j$, all $\mathbf{d}_j$ are real vectors, and $c_i$ and $e_j$ are real scalars.
The variable $\mathbf{x}$ is restricted to take only integer values.

Our optimization problem for finding class relationships (\ref{eq:objective_M}) equals to:
\begin{equation}
    \underset{\mathbf{M} \in \mathcal{M}}{\text{min}} \quad - \frac{1}{N} \text{tr}(\mathbf{Y}_{oh}^T \mathbf{P} \mathbf{M}) + \lambda_M \left( \frac{1}{K_C} \text{tr}(\mathbf{M}^T\boldsymbol{1}_{K_F}\boldsymbol{1}_{K_F}^T\mathbf{M}) -  \frac{K_F^2}{K_C^2} \right)
\end{equation}
with
\begin{equation}
       \mathcal{M} = \{ \mathbf{M} \in \{0, 1\}^{K_F \times K_C} \, | \,  \mathbf{M}\boldsymbol{1}_{K_C} = \boldsymbol{1}_{K_F}, \mathbf{M}^T\boldsymbol{1}_{K_F} \geq \boldsymbol{1}_{K_C} \}.
\end{equation}
We can rewrite our objective as:
\begin{equation}
    \underset{\mathbf{M} \in \mathcal{M}}{\text{min}} \quad \frac{\lambda_M}{K_c} \left( \text{tr}(\mathbf{M}^T\boldsymbol{1}_{K_F}\boldsymbol{1}_{K_F}^T\mathbf{M})\right) + \frac{-1}{N} \text{tr}(\mathbf{Y}_{oh}^T \mathbf{P} \mathbf{M})  + \left( - \frac{\lambda_M \cdot K_F^2}{K_C^2} \right).
\end{equation}
By defining $\mathbf{A}_0 = \frac{\lambda_M}{K_C} \boldsymbol{1}_{K_F}\boldsymbol{1}_{K_F}^T$, $\mathbf{B}_0 = \frac{-1}{N} \cdot \mathbf{Y}_{oh}^T \mathbf{P}$ and $c_0 =  - \frac{\lambda_M \cdot K_F^2}{K_C^3}$ we get:
\begin{equation}
\label{eq:qc_M}
    \underset{\mathbf{M}}{\text{min}} \quad \text{tr}(\mathbf{M}^T\mathbf{A}_0\mathbf{M} + \mathbf{B}_0 \mathbf{M}  + c_0 \cdot \mathbf{I}_{K_C}).
\end{equation}
$\mathbf{A}_0$ is a matrix of ones scaled by a positive coefficient $\frac{\lambda_M}{K_C}$ and therefore symmetric positive semidefinite, $\mathbf{I}_{K_C}$ is $K_C$-dimensional identity matrix, while $\mathbf{B}_0$ and $c_0$ are real. 
We now turn to the set of feasible solutions $\mathcal{M}$.
Let $\mathbf{M} \in \{0, 1\}^{K_F \times K_C}$ be a binary matrix, then we can recover $\mathcal{M}$ with the following constraints written in the standard form:
\begin{equation}
\label{eq:const_ineq}
     - \mathbf{M}^T\boldsymbol{1}_{K_F} + \boldsymbol{1}_{K_C} \leq \boldsymbol{0}_{K_C}
\end{equation}
\begin{equation}
\label{eq:const_eq}
    \mathbf{M}\boldsymbol{1}_{K_C} - \boldsymbol{1}_{K_F} = \boldsymbol{0}_{K_F}.
\end{equation}
 $\boldsymbol{1}_d$ is $d$-dimensional column vector of ones and $\boldsymbol{0}_d$ is $d$-dimensional column vector of zeros.
Thus, the objective (\ref{eq:qc_M}) with the constraints (\ref{eq:const_ineq}) and (\ref{eq:const_eq}) is integer quadratic program. 





\section{FALCON Algorithm for Multiple Datasets}
\label{appendix:multisource_alg}
\renewcommand{\thetable}{\Alph{section}\arabic{table}}
\renewcommand\thefigure{\Alph{section}\arabic{figure}} 
\renewcommand\thealgorithm{\Alph{section}\arabic{algorithm}}
\renewcommand{\theHtable}{\Alph{section}\arabic{table}}
\renewcommand\theHfigure{\Alph{section}\arabic{figure}} 
\renewcommand\theHalgorithm{\Alph{section}\arabic{algorithm}}
\setcounter{table}{0}
\setcounter{figure}{0}
\setcounter{algorithm}{0}

Algorithm \ref{alg:fg_multi} extends the Algorithm \ref{alg:fg_cls} to simultaneously train FALCON on multiple datasets.
Different than the standard algorithm, we now learn $D$ class relationships matrices $\mathbf{M}$.
Still, all of them can be learned in parallel.
Other algorithm parts remain the same.
The steps of the algorithm are outlined in Algorithm 2.
\begin{algorithm}[ht]
   \caption{FALCON training on multiple datasets}
   \label{alg:fg_multi}
\begin{algorithmic}
   \STATE {\bfseries Input:} Fine-grained classifier $f_\theta$, collection of D datasets $\mathcal{D}_\text{all}$, number of fine-grained classes $K_F$, hyperparameters $\boldsymbol{\lambda}, T, \lambda_M$, $\phi$
   \STATE $\theta = \text{selfsup\_initialization}(\mathcal{D})$
   \FOR{$l=1$ {\bfseries to} D}
   \STATE $\mathbf{M}_l$  = solve (\ref{eq:objective_M}) for random cost matrix
   \ENDFOR
   \FOR{$i=1$ {\bfseries to} max\_epochs}
   \FOR{$j=1$ {\bfseries to} max\_iters}
   \STATE Sample minibatch $(\mathbf{x}, y, l)$ from $\mathcal{D}_\text{all}$
   \STATE $L_\text{cls}$ =  evaluate loss (\ref{eq:joint_objective}) with $\mathbf{M} = \mathbf{M}_l$
    \STATE $L_\text{cons}$ =  evaluate loss (\ref{eq:cons}) with $\mathbf{M} = \mathbf{M}_l$
   \STATE $L_\text{reg}$ =  evaluate loss (\ref{eq:reg_ent}) over the whole minibatch
   \STATE $L = \lambda_1 \cdot L_\text{cls} + \lambda_2 \cdot L_\text{cons} + \lambda_3 \cdot L_\text{reg}$
   \STATE $\theta = \text{SGD}(L, \theta)$
   \IF{j \% $P$ = 0}
    \FOR{$l=1$ {\bfseries to} D}
    \STATE \textit{\# Note: all D subroutines can run in parallel}
   \STATE Gather $\mathbf{Y}_{oh}$ and $\mathbf{P}$ using $f_\theta$ over $\mathcal{D}_l$
   \STATE $\mathbf{C} = \mathbf{Y}_{oh}^T\mathbf{P}$
   \STATE $\mathbf{M}_l$ =  solve (\ref{eq:objective_M}) for cost matrix $\mathbf{C}$
   \ENDFOR
   \ENDIF
   \ENDFOR
   \ENDFOR
\STATE {\bfseries Output:} fine-grained classifier parameters $\theta$, set of class relationships $\mathbf{M}_1, \dots, \mathbf{M}_D$
\end{algorithmic}
\end{algorithm}

\section{Dataset Details}
\label{appendix:dataset_details}
\renewcommand{\thetable}{\Alph{section}\arabic{table}}
\renewcommand\thefigure{\Alph{section}\arabic{figure}} 
\renewcommand\thealgorithm{\Alph{section}\arabic{algorithm}}
\renewcommand{\theHtable}{\Alph{section}\arabic{table}}
\renewcommand\theHfigure{\Alph{section}\arabic{figure}} 
\renewcommand\theHalgorithm{\Alph{section}\arabic{algorithm}}
\setcounter{table}{0}
\setcounter{figure}{0}
\setcounter{algorithm}{0}

\textbf{Living17, Nonliving26, Entity13, Entity30} are four image datasets from BREEDS benchmark \cite{santurkar21iclr}.
Every dataset is a subset of ImageNet-1k.
Class relations are obtained by merging ImageNet synsets according to the WordNet \cite{miller94wordnet} 
lexical database.
We use the standard image resolution of $224 \times 224$.
All four datasets are balanced in terms of fine-grained classes associated with every coarse class and in terms of dataset samples associated with every fine class.

\textbf{CIFAR100} \cite{krizhevsky09cifar} is a well known dataset which contains small $32 \times 32$ images with the corresponding coarse and fine labels.
The dataset is balanced in terms of fine-grained classes associated with every coarse class and in terms of dataset samples associated with every fine-grained class.

\textbf{CIFAR68} is created from CIFAR100 by removing the following 32 fine-grained classes: 'apple', 'baby', 'beetle', 'bottle', 'boy', 'camel', 'can', 'chimpanzee', 'clock', 'couch', 'crocodile', 'crab', 'dolphin', 'lamp', 'leopard', 'lobster', 'maple', 'mountain', 'mouse', 'mushroom', 'pear', 'plate', 'rose', 'seal', 'streetcar', 'tank', 'tiger', 'tractor', 'train', 'turtle', 'wardrobe', and 'whale'.
These classes are selected randomly.
As a result, the dataset has an imbalanced number of fine-grained classes associated with every coarse class.

\begin{figure}[ht]
    \centering
    \includegraphics[width=\linewidth]{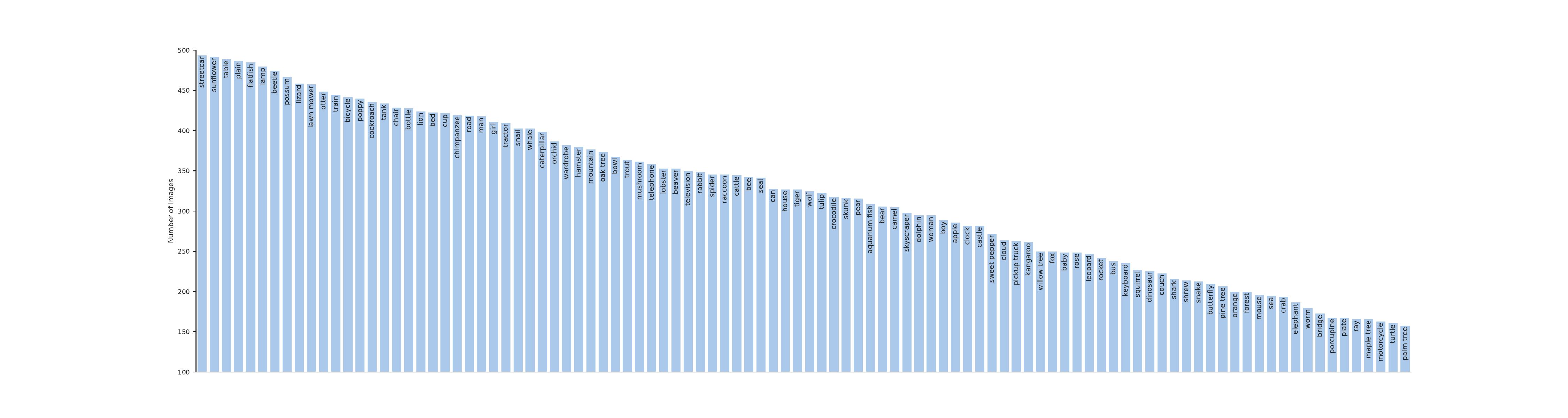}
    \vskip -0.1in
    \caption{Number of images in every fine-grained class of sample imbalanced version of the CIFAR100 dataset.}
    \label{fig:si_c100}
\end{figure}

\textbf{CIFAR-SI} is created from CIFAR100 by removing up to $70\%$ of training samples from every fine-grained class.
Thus we effectively disbalanced the number of samples within every fine as well as within every coarse class.
Figure \ref{fig:si_c100} shows the number of samples in every fine-grained class of the dataset.

\textbf{tieredImageNet} is a subset of ImageNet-1k with non overlapping train, val and test taxonomies \cite{ren18iclr}.
We merge all three taxonomies to recover 
fine training classes grouped into 34 coarse classes.
We combine the resulting taxonomy with the standard ImageNet splits.
Also, we preserve the original resolution of the images.
The resulting dataset has an imbalanced number of fine-grained classes associated with every coarse class.

\textbf{PBMC} \cite{lindeboom23medrxiv} is a dataset with single-cell RNA-seq data extracted from blood samples of COVID-19 patients.
The dataset contains samples labeled according to coarse-grained cell types and fine-grained cell states.
Cell states do not overlap between different cell types.
We predict cell states given the knowledge of cell type.
The datasets is imbalanced in terms of fine-grained classes associated with every coarse class and in terms of dataset samples associated with every fine-grained class.
\begin{table}[ht]
\setlength{\tabcolsep}{2pt}
\caption{Detailed overview of nine evaluation datasets.
}
\label{tab:dataset_info2}
\vskip 0.05in
\begin{center}
\begin{small}
\begin{sc}
\begin{tabular}{lccccccccc}
\toprule
Dataset & Living17 & Nonliving26 & Entity30 & Entity13 & CIFAR100 & CIFAR68 & CIFAR-SI & tieredIN & PBMC \\
\midrule
Coarse lasses & 17 & 26 & 30 & 13 & 20 & 20 & 20 & 34 & 27 \\
Fine classes & 68 & 104 & 240 & 260 & 100 & 68 & 100 & 608 &  83 \\
Resolution & $224^2$ & $224^2$ & $224^2$ & $224^2$ & $32^2$ & $32^2$ & $32^2$ & $224^2$ & 2k \\
Train samples & 88K & 132K & 307K & 334K & 45K & 30.6K & 29K & 779K & 372K \\
Val samples & 8.8K & 13K & 31K & 33K & 5K & 3.4K & 3.2K & - & -\\
Test samples& 3.4K & 5.2K & 12K & 13K & 10K & 6.8K & 6,4K & 30.4K  & - \\
Balanced samples & \cmark & \cmark & \cmark & \cmark & \cmark & \cmark & \xmark & \xmark & \xmark \\
Balanced classes & \cmark & \cmark & \cmark & \cmark & \cmark & \xmark & \cmark & \xmark & \xmark \\
\bottomrule
\end{tabular}
\end{sc}
\end{small}
\end{center}
\vskip -0.2in
\end{table}

\section{Implementation Details and Hyperparameter Search}
\label{appendix:implementation_details}
\renewcommand{\thetable}{\Alph{section}\arabic{table}}
\renewcommand\thefigure{\Alph{section}\arabic{figure}} 
\renewcommand\thealgorithm{\Alph{section}\arabic{algorithm}}
\renewcommand{\theHtable}{\Alph{section}\arabic{table}}
\renewcommand\theHfigure{\Alph{section}\arabic{figure}} 
\renewcommand\theHalgorithm{\Alph{section}\arabic{algorithm}}
\setcounter{table}{0}
\setcounter{figure}{0}
\setcounter{algorithm}{0}

In the case of large images, our fine-grained classifier is a ResNet50 \cite{he16cvpr} initialized with weights pre-trained in the self-supervised fashion on the ImageNet dataset \cite{chen21iccv}.
We train our method for 60 epochs with a batch size of 1024 across 2 GPUs for BREEDS datasets and for 90 epochs with batch size of 2048 across 4 GPUs for tieredImageNet.
We use SGD with a momentum of 0.9 and no weight decay.
The initial learning rate is set to 0.1 and annealed to 0.001 according to the cosine schedule with restarts every 30 epochs. 
This optimization procedure is similar to \citet{bukchin21cvpr,ni22iclr}.
We apply weak image augmentations to the input for $\theta_\text{EMA}$ and strong image augmentations for $\theta$, as in \citet{chen21iccv}.
In the case of small images, we use ResNet18 pre-trained in the self-supervised fashion on the CIFAR100 dataset, and train for 100 epochs with a batch size of 256.
Also, we decrease the learning rate by a factor of 10 after 60 and 80 epochs.
In the case of single-cell RNA transcriptomics, we use 4 layer MLP with 64 hidden units and ReLU activations.
We apply the model to 2k highly variable genes and do not apply any input augmentations.
We train for 10 epochs with a batch size of 1024 and decrease the learning rate by a factor of 10 after 6 and 8 epochs.
We find the nearest neighbours by measuring the distance in self-supervised representation space for images and in the raw input space for single-cell RNA transcriptomics.
We conducted all experiments using NVIDIA A100 GPUs. 

\begin{table}[ht]
\caption{The selected hyperparameters.}
\vskip 0.05in
\begin{center}
\begin{small}
\label{tab:hyperparams}
\begin{tabular}{cccc}
\toprule
Hyperparameter & Description & Value & Range \\
\midrule
$\lambda_1$ & Hyperparameter in (\ref{eq:final_cls}) & 0.5 & $\{ 0.5, 1.0, 1.5, 2.0\}$ \\
$\lambda_2$ & Hyperparameter in (\ref{eq:final_cls}) & 0.5 &  $\{ 0.5, 1.0, 1.5, 2.0\}$ \\
$\lambda_3$ & Hyperparameter in (\ref{eq:final_cls}) & 2 & $\{ 0, 1, 2, 3\}$ \\
$P$ 
& Number of iterations in Alg.\ \ref{alg:fg_cls} & 20 & $\{ 10, 20, 30, 40, 50 \}$  \\
$\lambda_M$ & Hyperparameter in (\ref{eq:objective_M}) &  0.1 & $\{ 1, 0.5, 0.1, 0.05, 0.01, 0.005, 0.001, 0.0005 \}$ \\
$T$ & Temperature in (\ref{eq:q_target}) & 0.9 & $\{ 0.5, 0.6, 0.7, 0.8, 0.9, 1, 1.1, 1.2, 1.3, 1.4, 1.5 \}$ \\
$\gamma$ & EMA parameter & 0.99 &  - \\
$L$ & Number of nearest neighbours in (\ref{eq:nn_loss}) & 20 & - \\
\bottomrule
\end{tabular}
\end{small}
\end{center}
\vskip -0.1in
\end{table}

The hyperparameter search was conducted using the TPE algorithm implemented with the Optuna framework \cite{akiba19kdd} on CIFAR100.
We ran 200 trials and evaluated performance after 30 epochs on a held-out validation set.
The range for every hyperparameter and the selected value are listed in the Table \ref{tab:hyperparams}.
The hyperparameters without range are fixed in the early stages of the research.
For example, we follow \citet{gansbeke20eccv} and use $L=20$.
We transfer the selected hyperparameters to all other datasets with the exception of $\lambda_3$ and $\lambda_M$.
$\lambda_3$ depends on the number of fine-grained classes hence we increase it for datasets with more fine-grained classes.
In the case of Entity13 and Entity30 we set it to 3 and in the case of tieredImageNet we set it to 5.
Similarly, we decrease $\lambda_M$ for the imbalanced datasets. 
We set it to $5\cdot10^{-2}$ in the case of CIFAR68 and to $5\cdot10^{-5}$ in the case of tieredImageNet.
These values are set without additional hyperparameter search.
We also transfer the majority of hyperparameters to single-cell datasets except for $\lambda_M$ which is set to $5 \cdot 10^{-3}$ and $\lambda_3$ which is set to 0.5.
We decrease these parameters due to imbalanced data and class distributions.

We solve discrete optimization problem using Gurobi solver \cite{gurobi23manual}.
Gurobi offers free licenses for academics.
We limit the runtime of the program to 30 seconds which is rarely reached.
Due to the number of classes in tieredImageNet, we increase the time limit to 120 seconds.
We compute the graph edit distance using \textit{GMatch4py}\footnote{\url{https://github.com/Jacobe2169/GMatch4py}}.

\section{Extended  Results }
\label{appendix:extended_results}
\renewcommand{\thetable}{\Alph{section}\arabic{table}}
\renewcommand\thefigure{\Alph{section}\arabic{figure}} 
\renewcommand\thealgorithm{\Alph{section}\arabic{algorithm}}
\renewcommand{\theHtable}{\Alph{section}\arabic{table}}
\renewcommand\theHfigure{\Alph{section}\arabic{figure}} 
\renewcommand\theHalgorithm{\Alph{section}\arabic{algorithm}}
\setcounter{table}{0}
\setcounter{figure}{0}
\setcounter{algorithm}{0}

Table \ref{tab:fine-grained-cls-std} presents the clustering accuracy (Acc) and adjusted rand index (ARI) for all the considered datasets. The number in subscript corresponds to the standard deviation over three runs.
\begin{table}[ht]
\setlength{\tabcolsep}{2pt}
\caption{Fine-grained clustering accuracy and ARI  performance on seven image datasets averaged over three runs. Standard deviation is reported as the subscript.}
\label{tab:fine-grained-cls-std}
\begin{center}
\begin{small}
\begin{sc}
\begin{tabular}{lcccccccccccccc}
\toprule
\multirow{2}{*}{Method} & \multicolumn{2}{c}{Living17} & \multicolumn{2}{c}{Nonliving26} & \multicolumn{2}{c}{Entity30} & \multicolumn{2}{c}{Entity13} & \multicolumn{2}{c}{CIFAR100} & \multicolumn{2}{c}{CIFAR68} & \multicolumn{2}{c}{tieredIN} \\
 & Acc & ARI & Acc & ARI & Acc & ARI & Acc & ARI & Acc & ARI & Acc & ARI & Acc & ARI \\
 \midrule
ERM & $86.3_{0.2}$ & $76.1_{0.3}$ & $84.6_{0.1}$ & $73.8_{0.2}$ & $85.6_{0.1}$ & $74.5_{0.1}$ & $85.9_{0.1}$ & $75.4_{0.1}$ & $74.5_{0.4}$ & $57.0_{0.6}$ & $78.8_{0.5}$ & $62.9_{0.8}$ & $79.1_{0.1}$ &  $65.1_{0.2}$\\[0.5em]
SCAN & $61.9_{0.1}$ & $50.1_{0.1}$ & $54.3_{0.1}$ & $39.7_{0.1}$ & $51.1_{0.1}$ & $38.4_{0.0}$ & $50.8_{0.0}$ & $37.5_{0.0}$ & $47.1_{0.1}$ & $34.4_{0.2}$ & $51.4_{0.3}$ & $39.8_{0.2}$ & $43.6_{0.0}$ & $28.9_{0.1}$ \\
ANCOR & $27.7_{0.4}$ & $36.1_{0.1}$ & $27.9_{0.6}$ & $34.7_{0.2}$ & $17.0_{0.2}$ & $20.2_{0.1}$ & $8.4_{0.4}$ & $8.5_{0.1}$ & $23.4_{0.6}$ & $26.6_{0.2}$ & $30.1_{0.5}$ & $33.7_{0.1}$ & $47.8_{0.1}$ & $34.1_{0.3}$ \\
SNCA & $39.2_{0.1}$ & $30.9_{0.6}$ & $43.6_{0.9}$ & $31.1_{1.0}$ & $36.1_{0.5}$ & $23.4_{0.2}$ & $35.1_{0.7}$ & $20.9_{0.3}$ & $42.9_{0.1}$ & $18.9_{1.3}$ & $47.6_{0.3}$ & $23.3_{3.2}$ & $22.3_{0.2}$ & $11.0_{0.1}$ \\
GEORGE &  $62.8_{1.5}$ & $53.2_{0.8}$ & $58.8_{1.0}$ & $47.2_{0.7}$ & $50.1_{0.4}$ & $35.6_{0.9}$ & $49.6_{0.3}$ & $35.7_{0.1}$ & $51.9_{0.3}$ & $36.0_{0.4}$ & $59.6_{0.9}$ & $42.8_{1.0}$ & $43.0_{0.2}$ & $29.1_{0.1}$\\
SCGM & $62.3_{3.0}$ & $49.3_{2.4}$ & $56.4_{0.1}$ & $42.0_{0.6}$ & $56.0_{0.1}$ & $41.4_{0.6}$ & $54.8_{0.4}$ & $40.8_{0.6}$ & $47.9_{2.8}$ & $32.2_{2.2}$ & $49.6_{1.4}$ & $34.7_{1.1}$ & $46.6_{0.6}$ &  $32.0_{0.5}$ \\
SCAN-C & $67.1_{0.1}$ & $54.7_{0.1}$ & $60.4_{0.1}$ & $45.8_{0.1}$ & $60.6_{0.1}$ & $46.2_{0.1}$ & $57.7_{0.1}$ & $43.7_{0.1}$ & $48.7_{0.2}$ & $36.1_{0.1}$ & $54.3_{0.2}$ & $41.9_{0.1}$ & $48.2_{0.1}$ & $33.2_{0.0}$ \\[0.5em]
FALCON & $\textbf{71.8}_{1.2}$ & $\textbf{60.3}_{1.5}$ & $\textbf{65.7}_{1.2}$ & $\textbf{55.5}_{0.5}$ & $\textbf{65.1}_{0.6}$ & $\textbf{53.3}_{0.5}$ & $\textbf{63.6}_{1.1}$ & $\textbf{51.9}_{0.5}$ & $\textbf{59.6}_{0.7}$ & $\textbf{42.5}_{0.3}$ & $\textbf{60.4}_{1.2}$ & $\textbf{47.0}_{1.3}$ & $\textbf{53.4}_{0.6}$ & $\textbf{41.6}_{0.2}$\\
\bottomrule
\end{tabular}
\end{sc}
\end{small}
\end{center}
\vskip -0.1in
\end{table}

Table \ref{tab:fine-grained-cls-rna} reports the standard deviation for the two datasets with the imbalanced number of samples in every fine-grained class. Additionally, we compute the accuracy score for each class independently and average the score over all classes.
The resulting macro averaged accuracy score is reported in the \textit{mAcc} column.
Macro accuracy provides a more balanced view of the classifier's performance on imbalanced datasets.

\begin{table*}[ht]
\setlength{\tabcolsep}{5pt}
\caption{Fine-grained performance on two datasets with an imbalanced number of samples in each fine-grained class. Results are averaged over three runs. }
\label{tab:fine-grained-cls-rna}
\vskip 0.1in
\begin{center}
\begin{small}
\begin{sc}
\begin{tabular}{lcccccc}
\toprule
\multirow{2}{*}{Method} & \multicolumn{3}{c}{$\text{CIFAR-SI}$} & \multicolumn{3}{c}{$\text{PBMC}$} \\
 & Acc & ARI & mAcc & Acc & ARI & mAcc \\
 \midrule
Upper Bound & $73.0 \pm 0.1$ & $56.6 \pm 0.2$ & $71.4 \pm 0.1 $ & $86.5 \pm 0.3$ & $83.7 \pm 0.1$ & $ 59.9 \pm 0.5 $ \\[0.5em]
SCAN & $47.4 \pm 0.3$ & $35.8 \pm 0.3$ & $44.9 \pm 0.4$ & $18.9 \pm 1.8$ & $12.8 \pm 1.7$  & $11.7 \pm 0.8$ \\
ANCOR & $28.7 \pm 0.6$ & $25.9 \pm 2.1$ & $24.7 \pm 0.1$ & $43.4 \pm 3.0$ & $35.3 \pm 2.8$ & $33.4 \pm 1.7$ \\
SNCA & $41.3 \pm 0.5$ & $21.6 \pm 0.6$ & $39.4 \pm 0.7$ &  $31.3 \pm 1.9$ & $23.0 \pm 2.3$ & $28.2 \pm 1.7$ \\
GEORGE &  $51.2 \pm 1.6$ & $36.7 \pm 1.7$ & $49.8 \pm 1.7$ &  $37.5 \pm 2.9$  & $34.5 \pm 4.5$ & $25.8 \pm 0.4$ \\
SCGM & $46.3 \pm 1.2$ & $31.8 \pm 0.7$ & $45.5 \pm 1.0$  & $23.9 \pm 1.9$ & $14.5 \pm 2.0$  & $21.6 \pm 1.7$ \\
SCAN-C &  $49.6 \pm 0.2$ & $38.0 \pm 0.1$ & $47.0 \pm 0.2$ &  $21.3 \pm 1.3 $ & $15.6 \pm 1.0$  & $11.2 \pm 1.2$  \\[0.5em]
FALCON &  $\textbf{55.6} \pm 0.2$ & $\textbf{39.1} \pm 1.1$ & $\textbf{53.4} \pm 0.5$  & $\textbf{74.2} \pm 0.5$ & $\textbf{71.9} \pm 0.4$ & $\textbf{35.5} \pm 0.9$ \\
\bottomrule
\end{tabular}
\end{sc}
\end{small}
\end{center}
\vskip -0.1in
\end{table*}


Table \ref{tab:abl_num_fine_grained_extended} shows the standard deviation for the graph edit distance.
\begin{table}[htb]
\setlength{\tabcolsep}{2pt}
\caption{Graph edit distance between the learnt class relations and the true class relations averaged over three runs.}
\label{tab:abl_num_fine_grained_extended}
\vskip 0.15in
\begin{center}
\begin{small}
\begin{sc}
\vskip -0.1in
\resizebox{\columnwidth}{!}{
\begin{tabular}{lccccccccc}
\toprule
GED & Living17 & Nonliving26 & Entity30 & Entity13 & CIFAR100 & CIFAR68 & CIFAR-SI &  tieredImageNet & PBMC \\
\midrule
SCGM & $30.7 \pm 6.1$ & $74.7 \pm 2.3$ & $98.7 \pm  10.1$ & $45.3 \pm 6.1$ & $61.3 \pm 12.2$ & $57.0 \pm 2.6 $ & $61.3 \pm 2.3$  & $132.0 \pm 0.0 $ &  $88.7 \pm 11.0 $ \\
FALCON & $\textbf{0.0} \pm 0.0$ & $\textbf{0.0} \pm 0.0$ & $\textbf{0.0} \pm 0.0$ & $\textbf{0.0} \pm 0.0$ & $\textbf{0.0} \pm 0.0$ & $\textbf{40.3} \pm 7.1 $ & $\textbf{0.0} \pm 0.0$ & $\textbf{110.7} \pm 9.2$ & $\textbf{73.3} \pm 2.1$  \\
\bottomrule
\end{tabular}
}
\end{sc}
\end{small}
\end{center}
\vskip -0.2in
\end{table}


Table \ref{tab:more_baselines} compares FALCON with two additional baselines derived from SCAN.
SCAN-per-coarse applies deep clustering within each coarse class independently and utilizes coarse classes for routing between different model instances.
SCAN-finetune utilizes SCAN-per-coarse to generate pseudo-labels for the training set.
The generated pseudo-labels are then utilized for supervised training of a single model. 
Both SCAN-per-coarse and SCAN-finetune perform worse than SCAN and SCAN-C that are considered in main experimental results.
\begin{table}[htb]
\setlength{\tabcolsep}{5pt}
\caption{Performance of FALCON with actual vs.~estimated class relationships.}
\label{tab:more_baselines}
\vskip 0.1in
\begin{center}
\begin{small}
\begin{sc}
\begin{tabular}{lcc}
\toprule
tieredImageNet & Acc & ARI \\
\midrule
SCAN-C & 48.2 & 33.2 \\
SCAN-per-coarse & 42.3 & 31.7 \\
SCAN-finetune & 33.4 & 17.9 \\
FALCON & 53.4 & 41.6\\
\bottomrule
\end{tabular}
\end{sc}
\end{small}
\end{center}
\vskip -0.2in
\end{table}

\section{Coarse classes of CIFAR100}
\label{appendix:cifar100_taxonomies}
\renewcommand{\thetable}{\Alph{section}\arabic{table}}
\renewcommand\thefigure{\Alph{section}\arabic{figure}} 
\renewcommand\thealgorithm{\Alph{section}\arabic{algorithm}}
\renewcommand{\theHtable}{\Alph{section}\arabic{table}}
\renewcommand\theHfigure{\Alph{section}\arabic{figure}} 
\renewcommand\theHalgorithm{\Alph{section}\arabic{algorithm}}
\setcounter{table}{0}
\setcounter{figure}{0}
\setcounter{algorithm}{0}

Table \ref{tab:taxonomies} lists coarse classes from the two considered taxonomies of CIFAR100.
We color the coarse classes shared between the two taxonomies in blue.
\begin{table}[ht]
\setlength{\tabcolsep}{4pt}
\caption{Different groupings of fine-grained classes into coarse classes.}
\label{tab:taxonomies}
\vskip 0.1in
\begin{center}
\begin{tiny}
\begin{tabular}{ccc|cc}
\toprule
\multirow{2}{*}{\#} & \multicolumn{2}{c|}{\textbf{Taxonomy T1}} & \multicolumn{2}{c}{\textbf{Taxonomy T2}} \\
 & \textbf{Coarse class} & \textbf{Fine-grained classes} & \textbf{Coarse class} & \textbf{Fine-grained classes} \\
 \midrule
 \multirow{1}{*}{1} &\multirow{1}{*}{Aquatic mammals} & beaver, dolphin, otter, seal, whale & \multirow{1}{*}{\color{blue}Trees} & maple, oak, palm, pine, willow \\\hline
\multirow{1}{*}{2} & \multirow{1}{*}{Fish} & aquarium fish, flatfish, ray, shark, trout & \multirow{1}{*}{\color{blue}Flowers} & orchid, poppy, rose, sunflower, tulip \\\hline
\multirow{1}{*}{3} & \multirow{1}{*}{\color{blue}Flowers} & orchid, poppy, rose, sunflower, tulip  & \multirow{1}{*}{\color{blue}Food containers} &  bottle, bowl, can, cup, plate \\ \hline
\multirow{1}{*}{4} & \multirow{1}{*}{\color{blue}Food containers} & bottle, bowl, can, cup, plate  & \multirow{1}{*}{\color{blue}Fruit and vegetables} &  apple, mushroom, orange, pear, sweet pepper  \\ \hline
\multirow{1}{*}{5} & \multirow{1}{*}{\color{blue}Fruit and vegetables} & apple, mushroom, orange, pear, sweet pepper & \multirow{1}{*}{\color{blue}Household electrical devices} & clock, keyboard, lamp, telephone, television \\\hline

\multirow{1}{*}{6} & \multirow{1}{*}{\color{blue}Household electrical devices} & clock, keyboard, lamp, telephone, television & \multirow{1}{*}{\color{blue}Household furniture} & bed, chair, couch, table, wardrobe \\\hline

\multirow{1}{*}{7} & \multirow{1}{*}{\color{blue}Household furniture} & bed, chair, couch, table, wardrobe & \multirow{1}{*}{Large carnivores} & bear, leopard, lion, tiger, wolf  \\\hline

\multirow{1}{*}{8} & \multirow{1}{*}{Insects} & bee, beetle, butterfly, caterpillar, cockroach & \multirow{1}{*}{Invertebrates} &  bee, beetle, butterfly,caterpillar, worm  \\\hline

\multirow{1}{*}{9} & \multirow{1}{*}{Large carnivores} & bear, leopard, lion, tiger, wolf & \multirow{1}{*}{Hard shelled animals} & crab, lobster, snail, turtle, cockroach \\\hline

\multirow{1}{*}{10} & Large man-made outdoor things  & bridge, castle, house, road, skyscraper  & \multirow{1}{*}{Small aquatic animals} & aquarium fish, flatfish,  ray, trout, otter \\ \hline

\multirow{1}{*}{11} & Large natural outdoor scenes   & cloud, forest, mountain, plain, sea & \multirow{1}{*}{Large aquatic animals} & beaver, dolphin, seal, crocodile, shark \\\hline

\multirow{1}{*}{12} & Large omnivores and herbivores  & camel, cattle, chimpanzee, elephant, kangaroo & \multirow{1}{*}{Outdoor scenes 2}  & cloud, sea, bridge, road, skyscraper \\\hline

\multirow{1}{*}{13} & \multirow{1}{*}{Medium-sized mammals} & fox, porcupine, possum, raccoon, skunk &  \multirow{1}{*}{Outdoor scenes 1} & forest, mountain,plain, castle, house   \\\hline

\multirow{1}{*}{14} & \multirow{1}{*}{Non-insect invertebrates} & crab, lobster,  snail, spider, worm  & \multirow{1}{*}{Large animals} & camel, cattle, elephant, whale, dinosaur  \\\hline

\multirow{1}{*}{15} & \multirow{1}{*}{People} & baby, boy,  girl, man, woman  & \multirow{1}{*}{Small mammals} &  shrew, squirrel, mouse, baby, raccoon\\\hline

\multirow{1}{*}{16} & \multirow{1}{*}{Reptiles} & crocodile, dinosaur, lizard, snake, turtle & \multirow{1}{*}{Medium sized mammals} & fox, porcupine, possum, skunk, kangaroo \\\hline

\multirow{1}{*}{17} & \multirow{1}{*}{Small mammals} & hamster, mouse, rabbit, shrew, squirrel & \multirow{1}{*}{Pets} & hamster, rabbit, lizard, snake, spider \\\hline

\multirow{1}{*}{18} & \multirow{1}{*}{\color{blue}Trees} & maple, oak, palm, pine, willow & \multirow{1}{*}{Primates} & chimpanzee, boy, girl,  man, woman \\\hline

\multirow{1}{*}{19} & \multirow{1}{*}{Vehicles 1} & bicycle, bus, motorcycle, pickup truck, train & \multirow{1}{*}{Personal vehicles} & bicycle, motorcycle, lawn mower, pickup truck, streetcar  \\\hline

\multirow{1}{*}{20} & \multirow{1}{*}{Vehicles 2} & lawn mower, rocket, streetcar, tank, tractor & \multirow{1}{*}{Transit vehicles} & bus, train, rocket, tank, tractor \\
\bottomrule
\end{tabular}
\end{tiny}
\end{center}
\end{table}

\section{Learning From Multiple Sources - Additional Experiments}
\label{appendix:multisource_additional_results}
\renewcommand{\thetable}{\Alph{section}\arabic{table}}
\renewcommand\thefigure{\Alph{section}\arabic{figure}} 
\renewcommand\thealgorithm{\Alph{section}\arabic{algorithm}}
\renewcommand{\theHtable}{\Alph{section}\arabic{table}}
\renewcommand\theHfigure{\Alph{section}\arabic{figure}} 
\renewcommand\theHalgorithm{\Alph{section}\arabic{algorithm}}
\setcounter{table}{0}
\setcounter{figure}{0}
\setcounter{algorithm}{0}

Table \ref{tab:multiple_sources2} shows fine-grained performance depending on the labeling policies.
The top row shows the fine-grained performance when CIFAR100 training images are labeled according to default coarse classes (T1).
The middle row shows the fine-grained performance when CIFAR100 training images are labeled according to our alternative coarse classes (T2).
In both cases, methods achieve comparable results which indicates that both coarse labeling policies are valid.
The last row shows the fine-grained performance when the half of samples are labeled according to T1  while the other half is labeled according to T2.
In all three cases, the training is conducted on the same number of samples. 
We observe performance improvements when trained on two coarse labelling policies for FALCON and only minor improvements for SCAN-C.
Overall, these results further strengthen our claim that FALCON exploits different coarse labels to learn better fine-grained predictions.
\begin{table}[ht]
\setlength{\tabcolsep}{5pt}
\caption{Impact of different labeling policies to fine-grained performance.}
\label{tab:multiple_sources2}
\begin{center}
\begin{small}
\begin{sc}
\begin{tabular}{cccccc}
\toprule
 \multirow{2}{*}{\# Samples} & \multirow{2}{*}{Taxonomy} & \multicolumn{2}{c}{SCAN-C} & \multicolumn{2}{c}{FALCON} \\
  &  & Acc & ARI & Acc & ARI \\
\midrule
N & T1 & 48.7 & 36.1 & 59.6 & 42.5  \\
 N & T2 & 49.0 & 36.3&  57.7 & 41.1 \\[.3em]
 N & T1\&T2 & 49.6 & 36.7 & \textbf{61.5} & \textbf{43.9} \\
\bottomrule
\end{tabular}
\end{sc}
\end{small}
\end{center}
\vskip -0.1in
\end{table}

\section{More Visual Examples}
\label{appendix:visual_examples_more}
\renewcommand{\thetable}{\Alph{section}\arabic{table}}
\renewcommand\thefigure{\Alph{section}\arabic{figure}} 
\renewcommand\thealgorithm{\Alph{section}\arabic{algorithm}}
\renewcommand{\theHtable}{\Alph{section}\arabic{table}}
\renewcommand\theHfigure{\Alph{section}\arabic{figure}} 
\renewcommand\theHalgorithm{\Alph{section}\arabic{algorithm}}
\setcounter{table}{0}
\setcounter{figure}{0}
\setcounter{algorithm}{0}

\begin{figure}[ht]
\vskip -0.1in
\begin{center}
\centerline{\includegraphics[width=\columnwidth]{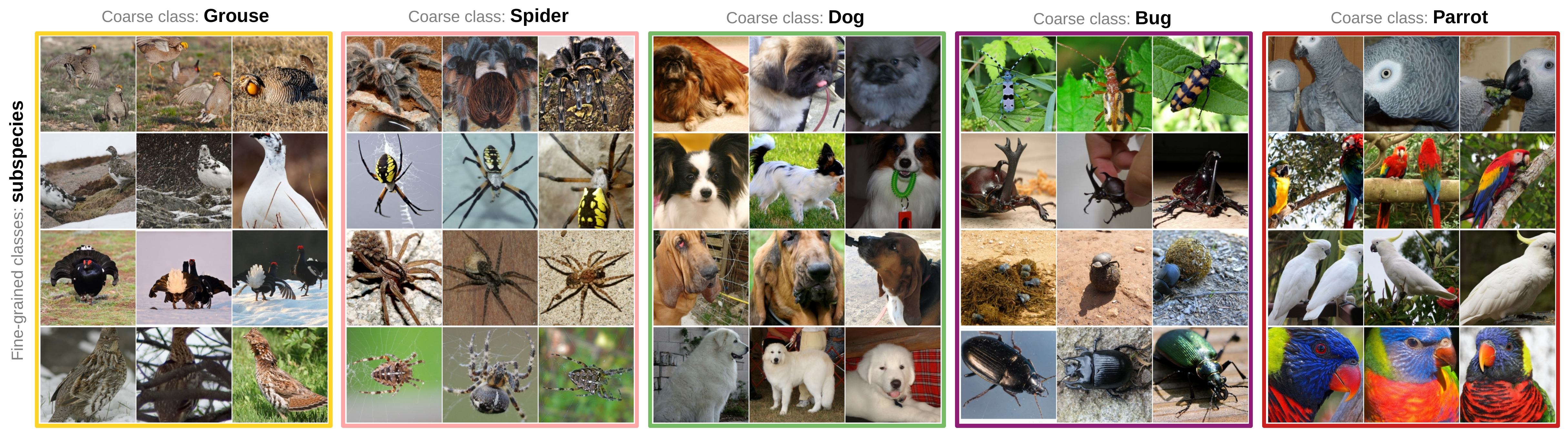}}
\caption{Three most confident predictions assigned to fine-grained classes. The fine-grained classes are grouped according to the learned class relationships.
The obtained subclasses correspond to subspecies.
}
\vskip -0.1in
\label{fig:cluster_visual_samples_more}
\end{center}
\end{figure}

Fig.\ \ref{fig:cluster_visual_samples_more} shows the three most confident predictions for different fine-grained classes associated with the same coarse class.
The fine-grained classes are grouped according to the learned class relationships.
The three samples in every fine-grained class indeed correspond to the same ground-truth fine classes.

\section{Subclasses of Fine-grained Classes in the Living17 Dataset}
\label{appendix:more_visual_subclasses_l17}
\renewcommand{\thetable}{\Alph{section}\arabic{table}}
\renewcommand\thefigure{\Alph{section}\arabic{figure}} 
\renewcommand\thealgorithm{\Alph{section}\arabic{algorithm}}
\renewcommand{\theHtable}{\Alph{section}\arabic{table}}
\renewcommand\theHfigure{\Alph{section}\arabic{figure}} 
\renewcommand\theHalgorithm{\Alph{section}\arabic{algorithm}}
\setcounter{table}{0}
\setcounter{figure}{0}
\setcounter{algorithm}{0}

\begin{figure}[ht]
\vskip -0.1in
\begin{center}
\centerline{\includegraphics[width=\columnwidth]{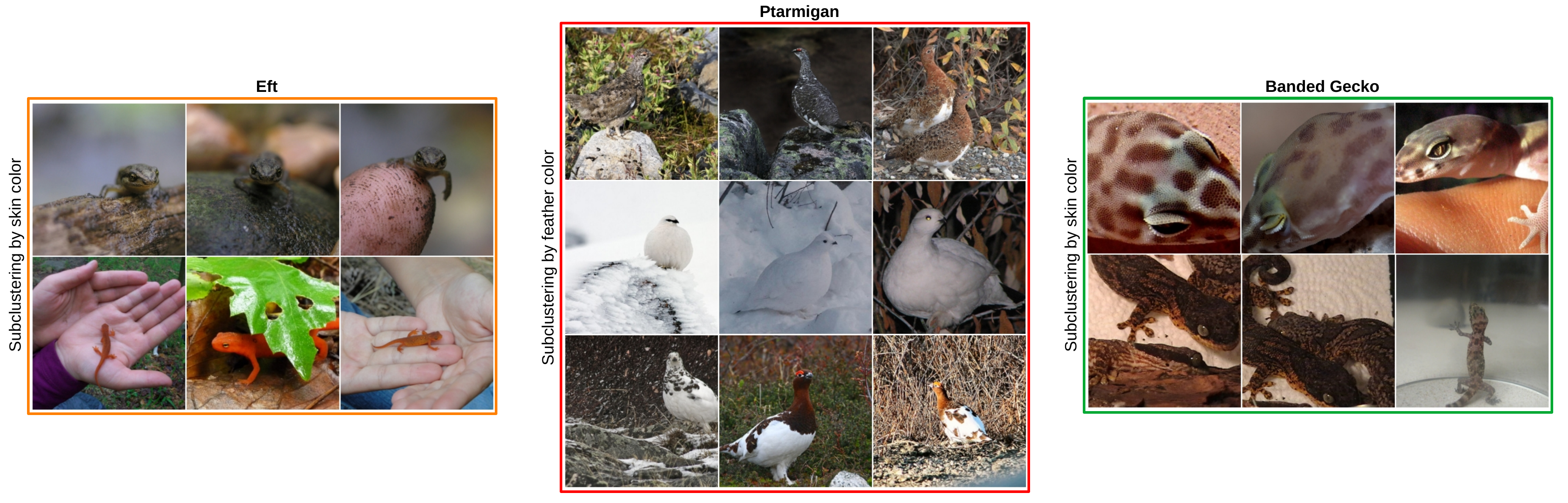}}
\caption{FALCON discovers subclasses within existing fine-grained classes, as shown for the three classes from the Living17 dataset.
The obtained subclasses correspond to subspecies.}
\vskip -0.1in
\label{fig:cluster_l17_samples_more}
\end{center}
\end{figure}

Figure \ref{fig:cluster_l17_samples_more} shows subclasses that FALCON discovered within the existing fine-grained classes of the Living17 dataset.
The discovered subclasses differ by skin or feather color, which indicates that they may correspond to different subspecies.

\section{Sensitivity on Number of Fine-grained Classes}
\label{appendix:sensitivity_on_number_of_fine}
\renewcommand{\thetable}{\Alph{section}\arabic{table}}
\renewcommand\thefigure{\Alph{section}\arabic{figure}} 
\renewcommand\thealgorithm{\Alph{section}\arabic{algorithm}}
\renewcommand{\theHtable}{\Alph{section}\arabic{table}}
\renewcommand\theHfigure{\Alph{section}\arabic{figure}} 
\renewcommand\theHalgorithm{\Alph{section}\arabic{algorithm}}
\setcounter{table}{0}
\setcounter{figure}{0}
\setcounter{algorithm}{0}

Table \ref{tab:sensitivity_k_f} analyzes FALCON performance for different numbers of fine-grained classes, controlled by hyperparameter $K_F$.
FLACON retains the majority of its performance for different values of $K_F$.
Most notably, FALCON preserves 95\% of its performance when $K_F$ is set to 1.5x the actual number of fine-grained classes.
\begin{table}[ht]
\setlength{\tabcolsep}{5pt}
\caption{FALCON performance for different numbers of fine-grained classes.}
\label{tab:sensitivity_k_f}
\vskip 0.15in
\begin{center}
\begin{small}
\begin{sc}
\begin{tabular}{lcccc}
\toprule
Accuracy & 80\% $K_F$ & $K_F$ & 120\% $K_F$ & 150\% $K_F$ \\
\midrule
FALCON & 55.1 & 60.4 & 57.5 & 57.6 \\
\bottomrule
\end{tabular}
\end{sc}
\end{small}
\end{center}
\vskip -0.1in
\end{table}

\section{Sensitivity to Loss Hyperparameters}
\label{appendix:hyperparameters_sensitiviy}
\renewcommand{\thetable}{\Alph{section}\arabic{table}}
\renewcommand\thefigure{\Alph{section}\arabic{figure}} 
\renewcommand\thealgorithm{\Alph{section}\arabic{algorithm}}
\renewcommand{\theHtable}{\Alph{section}\arabic{table}}
\renewcommand\theHfigure{\Alph{section}\arabic{figure}} 
\renewcommand\theHalgorithm{\Alph{section}\arabic{algorithm}}
\setcounter{table}{0}
\setcounter{figure}{0}
\setcounter{algorithm}{0}

We validate hyperparameter sensitivity on CIFAR100.
All results are averaged over three runs.
Table \ref{tab:abl_lambda_m} shows fine-grained accuracy, adjusted rand index, and graph edit distance depending on the value of $\lambda_M$ in (\ref{eq:objective_M}).
Our method keeps strong fine-grained performance even for suboptimal values of $\lambda_M$.
Decreasing the value of $\lambda_M$ reduces the influence of regularization (\ref{eq:M_bal}), thus enabling experimentation on imbalanced datasets.
Table \ref{tab:abl_T} shows fine-grained accuracy and adjusted rand index for different values of temperature hyperparameter in (\ref{eq:q_target}).
Our method keeps strong fine-grained performance for different values of $T$.
\begin{table}[ht]
  \begin{minipage}[b]{0.45\linewidth}
  \caption{Fine-grained performance for different $\lambda_M$ on CIFAR100.}
\label{tab:abl_lambda_m}
\begin{center}
\begin{small}
\begin{sc}
\begin{tabular}{cccc}
\toprule
$\lambda_M$ & Acc & ARI & GED \\
\midrule
0.5 & 59.3 & 42.2 & 0.0  \\
 \textbf{0.1} & 59.6 & 42.5 & 0.0 \\
0.05 & 59.0 & 41.9 & 0.0 \\
\bottomrule
\end{tabular}
\end{sc}
\end{small}
\end{center}
  \end{minipage}
  \hspace{0.05\linewidth}
  \begin{minipage}[b]{0.45\linewidth}
  \caption{Fine-grained performance  for different $T$ on CIFAR100.}
\label{tab:abl_T}
\begin{center}
\begin{small}
\begin{sc}
\begin{tabular}{ccc}
\toprule
$T$ & Acc & ARI\\
\midrule
0.6 & 57.1 & 40.5  \\
0.8 & 58.3 & 41.1  \\
\textbf{0.9} & 59.6 & 42.5  \\
\bottomrule
\end{tabular}
\end{sc}
\end{small}
\end{center}
  \end{minipage}
\end{table}

Tables \ref{tab:abl_l1}, \ref{tab:abl_l2} and \ref{tab:abl_l3} show fine-grained performance for different values of loss modulations hyperparameters $\lambda_1, \lambda_2,$ and $\lambda_3$ in (\ref{eq:final_cls}).
FALCON keeps strong performance for different values of these hyperparameters.
\begin{table}[ht]
  \begin{minipage}[b]{0.3\linewidth}
  \caption{Fine-grained performance for different $\lambda_1$ on CIFAR100.}
\label{tab:abl_l1}
    \begin{center}
\begin{small}
\begin{sc}
\begin{tabular}{ccc}
\toprule
$\lambda_1$ & Acc & ARI\\
\midrule
0.6 & 58.3 & 41.1  \\
\textbf{0.5} & 59.6 & 42.5  \\
0.4 & 58.6 & 42.0  \\
\bottomrule
\end{tabular}
\end{sc}
\end{small}
\end{center}
  \end{minipage}
  \hspace{0.05\linewidth}
  \begin{minipage}[b]{0.3\linewidth}
  \caption{Fine-grained performance for different $\lambda_2$ on CIFAR100.}
\label{tab:abl_l2}
    \begin{center}
\begin{small}
\begin{sc}
\begin{tabular}{ccc}
\toprule
$\lambda_2$ & Acc & ARI\\
\midrule
0.6 & 58.5  & 41.8  \\
\textbf{0.5} & 59.6 & 42.5  \\
0.4 & 58.6 & 41.8  \\

\bottomrule
\end{tabular}
\end{sc}
\end{small}
\end{center}
  \end{minipage}
  \hspace{0.05\linewidth}
  \begin{minipage}[b]{0.3\linewidth}
  \caption{Fine-grained performance for different $\lambda_3$ on CIFAR100.}
\label{tab:abl_l3}
    \begin{center}
\begin{small}
\begin{sc}
\begin{tabular}{ccc}
\toprule
$\lambda_3$ & Acc & ARI\\
\midrule
2.2 & 59.0 &  41.6 \\
\textbf{2.0} & 59.6 & 42.5  \\
1.8 & 59.1 & 42.0 \\

\bottomrule
\end{tabular}
\end{sc}
\end{small}
\end{center}
  \end{minipage}
\end{table}


\section{FALCON performance with existing class relations}
\label{appendix:estim_vs_actual}
\renewcommand{\thetable}{\Alph{section}\arabic{table}}
\renewcommand\thefigure{\Alph{section}\arabic{figure}} 
\renewcommand\thealgorithm{\Alph{section}\arabic{algorithm}}
\renewcommand{\theHtable}{\Alph{section}\arabic{table}}
\renewcommand\theHfigure{\Alph{section}\arabic{figure}} 
\renewcommand\theHalgorithm{\Alph{section}\arabic{algorithm}}
\setcounter{table}{0}
\setcounter{figure}{0}
\setcounter{algorithm}{0}

Table \ref{tab:actual_estimated_extended} compares the fine-grained classification performance of FALCON when class relations are estimated (as described in Section \ref{subsec:class_relationships}) against FALCON when class relations are available beforehand.
FALCON preserves the majority of its performance even when class relations are not available beforehand, indicating that it effectively infers the underlying class relations in most cases. 
\begin{table}[ht]
\setlength{\tabcolsep}{4pt}
\caption{FALCON performance when class relations are available beforehand vs.~FALCON performance when class relations are estimated (as described in Section \ref{subsec:class_relationships}).}
\label{tab:actual_estimated_extended}
\begin{center}
\begin{small}
\begin{sc}
\begin{tabular}{lccccccccccccccc}
\toprule
 & \multicolumn{3}{c}{CIFAR100} & \multicolumn{3}{c}{CIFAR68} & \multicolumn{3}{c}{Nonliving26} & \multicolumn{3}{c}{Entity30} & \multicolumn{3}{c}{tieredIN} \\
 & Acc & ARI & GED & Acc & ARI & GED & Acc & ARI & GED & Acc & ARI & GED & Acc & ARI & GED \\\midrule
Actual M & 59.6 & 42.5 & 0.0 & 67.2 & 50.7 & 0.0 & 67.8 & 56.1 & 0.0 & 67.5 & 54.4 & 0.0 & 53.9 & 41.7 & 0.0 \\
Estimated M & 59.6 & 42.5 & 0.0 & 60.4 & 47.0 & 40.3 & 65.7 & 55.5 & 0.0 & 65.1 & 53.3 & 0.0 & 53.4 & 41.6 &  110.7 \\\bottomrule
\end{tabular}
\end{sc}
\end{small}
\end{center}
\end{table}

\end{document}